\documentclass[conference]{IEEEtran}
\IEEEoverridecommandlockouts

\usepackage{cite}
\usepackage{amsmath,amssymb,amsfonts}
\usepackage{algorithmic}
\usepackage{graphicx}
\usepackage{textcomp}
\usepackage{xcolor}
\def\BibTeX{{\rm B\kern-.05em{\sc i\kern-.025em b}\kern-.08em
    T\kern-.1667em\lower.7ex\hbox{E}\kern-.125emX}}

\usepackage{microtype}

\usepackage{inconsolata}

\usepackage[ruled,vlined]{algorithm2e}
\usepackage{adjustbox}
\usepackage{booktabs}
\usepackage{caption}
\usepackage{enumitem}
\usepackage{float}
\usepackage{longtable}
\usepackage{linguex}
\usepackage{listings}
\usepackage{multirow}
\usepackage{soul}
\usepackage{subcaption}
\usepackage{relsize}
\usepackage{tabularray}
\usepackage{tikz}
\usepackage{comment}
\usepackage{url}

\begin{document}

\title{Emotion Across Speech and Faces: Shared Affective Mechanisms in Multimodal Foundation Models}

\author{
 \textbf{Xiutian Zhao\textsuperscript{1}},
 \textbf{Luqi Sun\textsuperscript{1}},
 \textbf{Bj\"orn Schuller\textsuperscript{2}}, 
 \textbf{Berrak Sisman\textsuperscript{1}}
\\
 \textsuperscript{1} Center for Language and Speech Processing (CLSP), Johns Hopkins University, USA \\
 \textsuperscript{2} Group on Language, Audio \& Music (GLAM), Imperial College London, UK
 }
\maketitle
\begin{abstract}
Modern multimodal foundation models (MFMs) have made rapid progress on tasks requiring integrated perception across speech, vision, and language, including emotion recognition. However, it remains unclear whether they recognize speech and facial emotion through shared affective functional units or modality-specific pathways. We explore emotion-sensitive neurons (ESNs), sparse decoder neurons selectively associated with emotion categories, in three MFMs: Gemma-4-12B-it, MiniCPM-o-4.5, and Qwen2.5-Omni-7B. Using speech emotion recognition and facial expression recognition as complementary probes, we identify acoustic and visual ESNs. Visual ESNs are causally meaningful: deactivating them selectively impairs recognition of the associated facial emotion, whereas steering their activations selectively enhances recognition of that emotion relative to other emotion categories. Acoustic and visual ESNs further show emotion-matched overlap and similar layer-wise distributions, indicating partial structural alignment between affective representations across speech and faces. Finally, cross-modal interventions reveal bidirectional causal transfer: ESNs identified from one modality produce emotion-specific effects when applied to the other. Our findings provide one of the first cross-modality activation-level analyses of affective functional units in MFMs, suggesting that speech and facial emotion recognition partially converge onto sparse decoder-level components that can be localized and manipulated without training.
\end{abstract}

\begin{IEEEkeywords}
multimodal foundation models, affective computing, speech emotion recognition, facial expression recognition, activation intervention
\end{IEEEkeywords}

\section{Introduction}

Emotion perception in natural communication is inherently multimodal. In face-to-face interaction, people infer affect from facial expressions, speech prosody, lexical content, body movement, and context \cite{10802538, 8070966, 10.1145/1027933.1027968, abdullah2021multimodal,9414654,e25101440,yang-etal-2025-towards-holistic}. Among these channels, speech and facial behavior are among the most commonly studied and practically important sources of affective information. Facial expressions provide visible cues such as muscle configuration and gaze, while speech conveys affect through pitch, rhythm, intensity, and timing \cite{wilce2009language, schirmer2017emotion, barrett2019emotional}. Psychological and affective computing studies have long debated the extent to which emotion perception reflects modality-general affective categories, modality-specific cue patterns, and contextual or culturally learned regularities \cite{russell1991culture}. This makes speech and facial emotion recognition a natural testbed for asking whether modern multimodal foundation models (MFMs) develop shared internal mechanisms for affect.

Recent progress in MFMs has substantially expanded the scope of machine perception. Models such as MiniCPM-o-4.5 \cite{yao2024minicpmvgpt4vlevelmllm}, Qwen2.5-Omni \cite{xu2025qwen25omnitechnicalreport}, and Gemma-4 \cite{gemma4_hf} can process images, speech, and text within a unified interface. These systems have achieved strong performance on audio and vision understanding tasks \cite{NEURIPS2023_3a2e5889,9721159,10448257,sakshi2025mmau,10889251,hacioglu-etal-2025-speechllms,chen-etal-2026-audio}, and have shown growing potential for affective applications, including speech emotion recognition (SER) and facial emotion recognition (FER) \cite{cheng2024emotionllama,10581982,Xu_Chen_Yu_Huang_Wu_Zhang_Li_Luo_Gu_2024,he2025meralionaudiollmbridgingaudiolanguage,li-etal-2025-emo-rl}. However, behavioral success alone does not reveal whether a model recognizes anger in a face and anger in a voice through related internal components, or whether cross-modal consistency emerges only near final answer selection.

Mechanistic interpretability  provides tools for localizing and manipulating internal components that support model behavior on affective tasks. Prior work has shown that individual neurons or sparse groups of units can align with human-interpretable concepts in vision models \cite{bau2017network,bau2020understanding,10.1609/aaai.v33i01.33016309}, language models \cite{voita-etal-2024-neurons,yu-ananiadou-2024-neuron,huben2024sparse}, and multimodal systems \cite{huang2024minerminingunderlyingpattern,fang2024towards,huo-etal-2024-mmneuron,xu2025deciphering}. In this work, a ``neuron'' refers to a scalar hidden unit inside a model component, and we focus on decoder MLP neurons because they provide a large set of nonlinear units through which multimodal representations are transformed before text generation \cite{gurnee2024universal,bau2019identifying,tang-etal-2024-language}. In parallel, affective modeling has studied how emotional style can be represented and controlled in speech generation and conversion systems \cite{wu2019end,xie2025emosteerttsfinegrainedtrainingfreeemotioncontrollable}. More recent work connects affect and interpretability by identifying emotion-related neurons or circuits in language and audio-language models, and by showing that activation steering can causally modulate emotional behavior \cite{lee-etal-2025-large,wang2025llmsfeelemotioncircuits,sofroniew2026twheemotion}. However, these analyses remain largely confined to a single modality, especially text or speech, leaving open whether affect-sensitive functional units are modality-specific or partially shared across perceptual channels.

This work investigates cross-modal emotion-sensitive neurons (ESNs) in MFMs. Following prior work on speech emotion recognition in large audio-language models \cite{zhao-etal-2026-discovering}, we define \emph{ESNs} as sparse decoder MLP units whose activation patterns are selectively associated with affective input categories and whose behavioral role can be tested through intervention. We distinguish \emph{acoustic emotion-sensitive neurons} (A-ESNs), identified from SER activations, from \emph{visual emotion-sensitive neurons} (V-ESNs), identified from FER activations. Across Gemma-4-12B-it \cite{gemma4_hf}, MiniCPM-o-4.5 \cite{yao2024minicpmvgpt4vlevelmllm}, and Qwen2.5-Omni-7B \cite{xu2025qwen25omnitechnicalreport}, we identify ESNs using a contrastive activation criterion, analyze their structural alignment through overlap and layer-wise distributions, and evaluate their causal role through deactivation and steering. Deactivation tests whether suppressing a selected neuron set disrupts recognition of the associated emotion, while steering tests whether amplifying that set selectively enhances recognition of the corresponding emotion.

Our first question is whether visual ESNs exist and whether they are functionally meaningful. Prior work has shown that SER-derived A-ESNs can be identified and causally validated in audio-language models \cite{zhao-etal-2026-discovering}. However, it remains unclear whether neurons with analogous functions can be discovered from affective tasks in other modalities, particularly vision. We therefore identify V-ESNs from correctly recognized FER examples and intervene on them during FER. The results show that V-ESNs are causally influential for MFMs' affective recognition behavior in a manner analogous to A-ESNs: deactivation selectively reduces recognition of the associated facial emotion, whereas steering increases predictions of the targeted emotion. These findings provide positive evidence that FER-derived V-ESNs constitute functional affective units, and show that ESNs can be independently discovered from both speech and facial emotion recognition tasks.

These observations motivate our next question: whether speech and facial emotion recognition converge onto related affective representations. To address this, we analyze the structural agreement between acoustic and visual ESNs inside the decoder. Specifically, we compare A-ESN and V-ESN sets using emotion-specific neuron-level overlap and layer-wise ESN distributions. Across models, the overlap is sparse but consistently stronger for matched emotion categories than for mismatched ones, suggesting partial cross-modal alignment. The layer-wise analysis further shows that both A-ESNs and V-ESNs are distributed across multiple decoder layers, with a tendency to concentrate more in middle and later layers than in early layers. This indicates that the observed ESNs are not isolated artifacts of a single block. Rather, affect-sensitive units form sparse depth-wise patterns that are broadly comparable across acoustic and visual inputs, although their precise localization remains model-dependent.

Structural overlap alone, however, does not establish functional sharing. Overlap between A-ESNs and V-ESNs could arise from incidental co-activation, shared answer-token processing, or biases introduced by the selection criterion. We therefore go beyond descriptive analysis and test whether the observed alignment is functional through cross-modal causal interventions. We transfer ESN masks across modalities: A-ESNs identified from speech are applied during FER, and V-ESNs identified from faces are applied during SER. These transferred interventions produce emotion-specific effects in both directions. In aggregate, deactivating transferred ESNs impairs recognition of the matched emotion more than non-matched emotions, while steering them increases the corresponding target-emotion response relative to non-target responses. These effects are consistently stronger and more structured than those produced by random controls. Thus, speech and facial emotion recognition in MFMs is not entirely modality-isolated; decoder MLPs contain sparse affective components that are partially transferable across speech and faces.

In summary, this study makes the following contributions. First, we identify and causally validate visual ESNs in MFMs, extending neuron-level affective analysis from speech to facial expression recognition. Second, we show that acoustic and visual ESNs exhibit sparse emotion-matched overlap and broadly similar layer-wise distributions. Third, \textbf{we present, to our knowledge, one of the first cross-modality analyses of affective functional units in MFMs, with a focus on speech and facial emotion recognition.} Our findings demonstrate bidirectional cross-modal causal transfer between speech- and face-derived ESNs, providing evidence that affective processing in MFMs partially relies on shared decoder-level components.

\section{Related Work}
Neuron-level specialization has been widely studied as a route to interpretable transformer-based models, with evidence that individual units or sparse unit groups can encode human-interpretable concepts in vision models \cite{bau2017network,bau2020understanding}, language models \cite{huben2024sparse,voita-etal-2024-neurons,yu-ananiadou-2024-neuron}, and multimodal systems \cite{kumar21d_interspeech,huang2024minerminingunderlyingpattern,fang2024towards,xu2025deciphering,huo-etal-2024-mmneuron}. In affective modeling, related work has examined how emotional and paralinguistic attributes are represented or controlled in speech systems, including learned style embeddings \cite{wu2019end}, continuous affect-control variables \cite{xie2025emosteerttsfinegrainedtrainingfreeemotioncontrollable}, and activation-based emotion steering or circuit analysis in language models \cite{konen-etal-2024-style, lee-etal-2025-large,wang2025llmsfeelemotioncircuits,sofroniew2026twheemotion}. These studies establish that affective behavior can often be localized or modulated through internal representations, but they do not directly address whether such representations are shared across perceptual modalities. Indeed, related evidence suggests that emotion-related neurons may not always transfer across input conditions, such as across languages in multilingual encoders \cite{singh-etal-2026-lost}.

Beyond text, probing and dissection methods have analyzed what audio and multimodal models encode internally, including phonetic, speaker, prosodic, and broader acoustic concepts \cite{singla2022audio,Akman2025Improving,yang2025audiolenscloserlookauditory,wu2024andaudionetworkdissection}. Most directly related to our work, recent studies have identified ESNs in large audio-language models for SER \cite{zhao-etal-2026-discovering, zhao2026multilingualemotionneuronslarge} and emotional voice conversion \cite{zhao2026neuronlevelemotioncontrolspeechgenerative}. However, existing analyses primarily focus on a single modality. It remains unclear whether ESNs are modality-specific or whether they form a partially shared substrate across different affective channels. Our work narrows this gap by identifying visual ESNs from facial expression recognition and testing their overlap and transfer with acoustic ESNs from SER in MFMs.

\section{Method}
\label{sec:method}

We apply a shared activation-based probing pipeline to SER and FER: collect decoder MLP activations on correctly recognized examples, select sparse ESNs using a contrastive criterion, and evaluate their causal role through deactivation and steering.

\subsection{Emotion-Conditioned Activation Collection}
\label{subsec:logging}

Let $\mathcal{E}$ denote the set of emotion categories. For each task modality $q \in \{\mathrm{SER}, \mathrm{FER}\}$ and emotion $e \in \mathcal{E}$, we run the original, unintervened model on the corresponding emotion recognition task. We use a multiple-choice question answering protocol and retain only correctly recognized examples for neuron identification, which reduces noise from model failure cases and yields cleaner emotion-conditioned activation statistics. Dataset-specific sample sizes for ESN identification are reported in Section~\ref{sec:setup}.

We instrument the decoder MLP modules and record the activated SwiGLU gate outputs \cite{shazeer2020gluvariantsimprovetransformer}. For layer $l$, neuron $n$, and valid token position $t$, let $a^{(q,e)}_{l,n,t}$ be the scalar gate activation under task modality $q$ and emotion $e$. We use a binary mask $m_t \in \{0,1\}$ to exclude irrelevant positions such as padding and instruction-only tokens. For each $(q,e)$, we compute the positive firing count and the number of valid positions as $K^{(q,e)}_{l,n}=\sum_t m_t \mathbb{I}(a^{(q,e)}_{l,n,t}>0)$ and $T^{(q,e)}=\sum_t m_t$. The normalized activation probability is then $P^{(q,e)}_{l,n}=K^{(q,e)}_{l,n}/T^{(q,e)}$, which measures how frequently a neuron fires for a given emotion and modality.

\subsection{Emotion-Sensitive Neuron Identification}
\label{subsec:selector}

We identify ESNs using Contrastive Activation Margin (ConAct) \cite{zhao-etal-2026-finding}. For each modality $q$, layer $l$, and neuron $n$, ConAct assigns the neuron to the emotion for which it fires most frequently and scores it by the margin between the highest and second-highest activation probabilities. Specifically, we define $e^{(1)}_{l,n}(q)=\arg\max_{e\in\mathcal{E}} P^{(q,e)}_{l,n}$, $P^{(1)}_{l,n}(q)=\max_{e\in\mathcal{E}}P^{(q,e)}_{l,n}$, and $P^{(2)}_{l,n}(q)=\max_{e\neq e^{(1)}_{l,n}(q)}P^{(q,e)}_{l,n}$. The emotion-specific ConAct score is
$s^{(q,e)}_{l,n}=P^{(1)}_{l,n}(q)-P^{(2)}_{l,n}(q)$ if $e=e^{(1)}_{l,n}(q)$, and $s^{(q,e)}_{l,n}=0$ otherwise.

For each modality-emotion pair $(q,e)$, we rank all decoder MLP neurons by $s^{(q,e)}_{l,n}$ and select the top fraction $r$ as the corresponding ESN set. In the FER experiments, we use $r=0.5\%$ of all decoder MLP neurons to obtain V-ESNs; the same selection strategy is used for A-ESNs in SER. We denote the selected neuron set for emotion $e$ as $\mathcal{I}^{(q,e)}$, with $\mathcal{I}^{(\mathrm{SER},e)}$ corresponding to A-ESNs and $\mathcal{I}^{(\mathrm{FER},e)}$ corresponding to V-ESNs. As a control, we also construct random
masks by selecting the same number of neurons uniformly from the decoder MLPs, without using activation statistics. We average results over five independent random masks.

\subsection{Causal Interventions: Deactivation and Steering}
\label{subsec:intervention}

Given an ESN set $\mathcal{I}^{(q,e)}$, we evaluate its causal role through two interventions applied to decoder MLP gate activations: deactivation and steering. Let $g_{l,t}\in\mathbb{R}^{D_l}$ be the activated gate vector at layer $l$ and token position $t$, where $g_{l,t}=\mathrm{act}(\mathrm{gate\_proj}(x_{l,t}))$. The modified gate vector is then used in the standard SwiGLU computation, i.e., $\mathrm{down\_proj}(\tilde{g}_{l,t}\odot \mathrm{up\_proj}(x_{l,t}))$.

\paragraph{Deactivation}
For deactivation, we suppress the selected neurons by setting their gate activations to zero. For neuron $n$ in layer $l$, the mask is $r_{l,n}=0$ if $n\in\mathcal{I}^{(q,e)}$ and $r_{l,n}=1$ otherwise, yielding $\tilde{g}^{\mathrm{deact}}_{l,t}=g_{l,t}\odot r_l$. 

\paragraph{Steering}
For steering, we increase the selected neurons' activations with a multiplicative gain. The steering mask is $s_{l,n}(\alpha)=1+\alpha$ if $n\in\mathcal{I}^{(q,e)}$ and $s_{l,n}(\alpha)=1$ otherwise, giving $\tilde{g}^{\mathrm{steer}}_{l,t}=g_{l,t}\odot s_l(\alpha)$. Unless otherwise stated, we set $\alpha=0.5$.

These interventions provide complementary loss- and gain-of-function tests. We interpret an ESN set as causally emotion-relevant when its effects are emotion-selective and stronger than same-size random-mask controls. The same protocol is used for both within-modality validation and cross-modal transfer.

\section{Experiment Setup}
\label{sec:setup}

\subsection{Datasets and Models}
We evaluate emotion recognition from both acoustic and visual inputs. For SER, we use MSP-Podcast \cite{11457331}; for FER, we use AffectNet \cite{8013713}. We focus on five emotion categories shared by the two tasks: anger, fear, happiness, neutral, and sadness. 
We sample 150 utterances per emotion for evaluation in SER task and 300 facial images per emotion for evaluation in FER. 
The remaining correctly recognized SER/FER examples are used for A-ESN/V-ESN identification, from which we sample 100 successful cases per emotion for both tasks. 
This separation ensures that neuron selection and causal evaluation are performed on disjoint instances.

We evaluate three open-source MFMs: Gemma-4-12B-it \cite{gemma4_hf}, MiniCPM-o-4.5 \cite{yao2024minicpmvgpt4vlevelmllm}, and Qwen2.5-Omni-7B \cite{xu2025qwen25omnitechnicalreport}. These models support both acoustic and visual inputs and expose decoder MLP modules for activation-level intervention. All models are evaluated under the same multiple-choice question answering protocol. To reduce position and label bias \cite{zhao-etal-2024-measuring}, the order of emotion options is randomized across examples, and the model is instructed to output only the option index (alphabetic letters) rather than the emotion word.

\subsection{Prompting and Decoding}
\label{sec:decoding}
All inference is performed with deterministic decoding, using greedy search with temperature 0. We set the maximum generation length to 20 tokens and apply lightweight post-processing to extract the predicted option index from the model output. 
Randomness enters through dataset sampling, option-order randomization, and random-mask construction; dataset sampling and option ordering are controlled by fixed seeds unless otherwise specified. For random-selection controls, we report averages over five independently sampled masks.

\subsection{Neuron Selection and Intervention Settings}
Unless stated otherwise, ESNs are selected with ConAct from decoder MLP activations. 
For each emotion, we select the top $r = 0.5\%$ of all decoder MLP neurons. In steering experiments, we use a multiplicative gain with $\alpha = 0.5$. 

\begin{table*}[ht!]
\centering
\resizebox{0.95\textwidth}{!}{%
\begin{tabular}{@{}cccccccc@{}}
\toprule
Evaluation Setting & MFM & Intervention & Unmasked & Random ($\Delta$) & Self-Emotion ($\Delta$) & Avg.\ Cross-Emotion ($\Delta$) & Self-Cross Gap \\ 
\midrule
                   & Gemma-4-  & Deactivation $\downarrow$ & \multirow{2}{*}{60.47} & 60.19 (-0.28)     & 34.67 (-25.80)          & 63.00 (+2.53)                 & -28.33         \\
                   & 12B-it    & Steering $\uparrow$       &                        & 60.13 (-0.34)     & 76.93 (+16.46)          & 55.92 (-4.55)                 & +21.01         \\ 
\cmidrule(l){2-8} 
\textbf{FER}       & MiniCPM-  & Deactivation $\downarrow$ & \multirow{2}{*}{64.00} & 64.13 (+0.13)     & 53.67 (-10.33)          & 64.83 (+0.83)                 & -11.16         \\
with V-ESN masks   & o-4.5     & Steering $\uparrow$       &                        & 63.99 (-0.01)     & 68.07 (+4.07)           & 63.27 (-0.73)                 & +4.80          \\ 
\cmidrule(l){2-8} 
                   & Qwen2.5-  & Deactivation $\downarrow$ & \multirow{2}{*}{48.60} & 47.53 (-1.07)     & 38.20 (-10.40)           & 48.85 (+0.25)                 & -10.65          \\
                   & Omni-7B   & Steering $\uparrow$       &                        & 47.76 (-0.84)     & 51.60 (+3.00)           & 47.72 (-0.88)                 & +3.88          \\ 
\midrule
                   & Gemma-4-  & Deactivation $\downarrow$ & \multirow{2}{*}{35.87} & 35.47 (-0.40)     & 20.93 (-14.94)          & 37.23 (+1.36)                 & -16.30         \\
                   & 12B-it    & Steering $\uparrow$       &                        & 35.65 (-0.22)     & 46.40 (+10.53)           & 34.13 (-1.74)                 & +12.27         \\ 
\cmidrule(l){2-8} 
\textbf{SER}       & MiniCPM-  & Deactivation $\downarrow$ & \multirow{2}{*}{27.47} & 27.01 (-0.46)      & 22.13 (-5.34)           & 29.13 (+1.66)                 & -7.00          \\
with A-ESN masks   & o-4.5     & Steering $\uparrow$       &                        & 27.33 (-0.14)     & 31.33 (+3.86)           & 26.67 (-0.80)                 & +4.66          \\ 
\cmidrule(l){2-8} 
                   & Qwen2.5-  & Deactivation $\downarrow$ & \multirow{2}{*}{36.93} & 36.99 (+0.06)     & 25.20 (-11.73)          & 38.67 (+1.74)                 & -13.47         \\
                   & Omni-7B   & Steering $\uparrow$       &                        & 36.83 (-0.10)     & 40.67 (+3.74)           & 36.07 (-0.86)                 & +4.60          \\ 
\bottomrule
\end{tabular}%
}
\caption{\textbf{Mono-modal causal effects of ESN interventions.}
V-ESN masks are identified and evaluated on FER; A-ESN masks are identified and evaluated on SER.
Columns report the unmasked UAR, random-mask controls, matched-emotion performance, average non-matched-emotion performance, and the Self-Cross Gap, as defined in Section~\ref{sec:eval_metrics}. $\Delta$ values indicate the changes relative to the unmasked UAR. Arrows ($\downarrow$, $\uparrow$) indicate the expected direction of the matched-emotion performance change.}
\label{tab:mono_results}
\end{table*}

\begin{figure*}[ht!]
    \centering
    \begin{subfigure}[b]{0.165\linewidth}
        \centering
        \includegraphics[width=\linewidth]{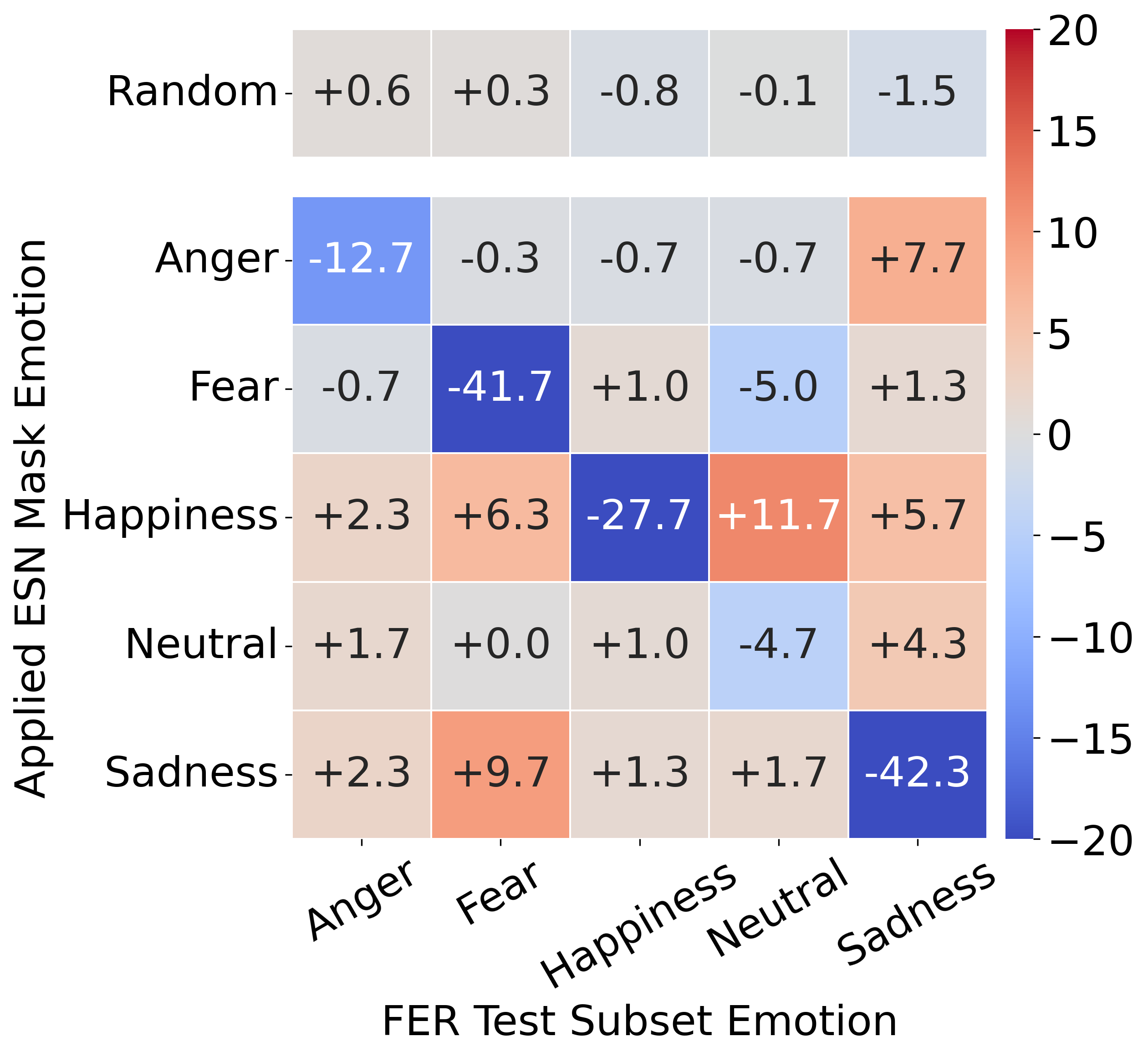}
        \caption{Deactivation,\\Gemma-4-12B-it}
    \end{subfigure}\hfill
    \begin{subfigure}[b]{0.165\linewidth}
        \centering
        \includegraphics[width=\textwidth]{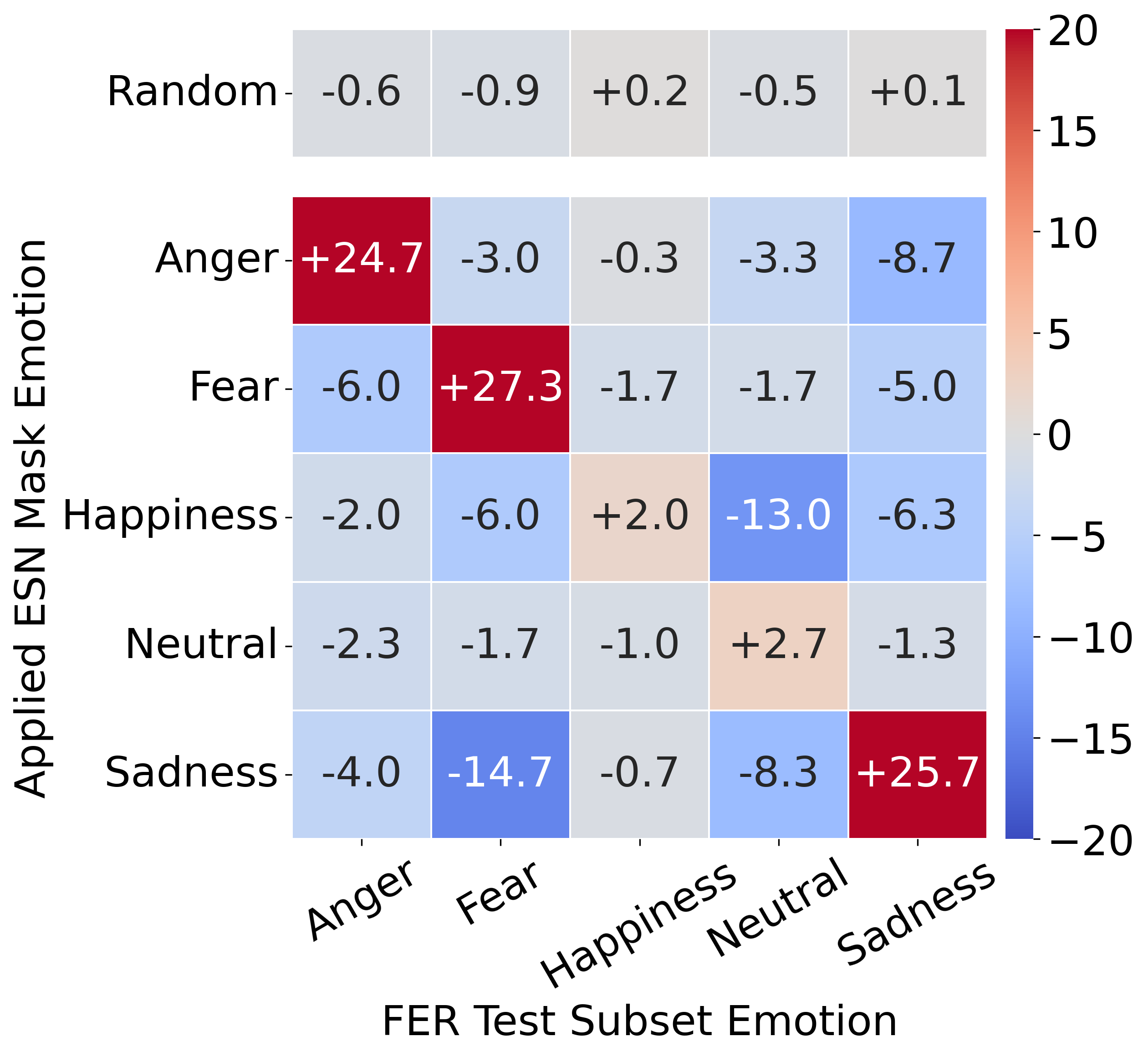}
        \caption{Steering,\\Gemma-4-12B-it}
    \end{subfigure}\hfill
    \begin{subfigure}[b]{0.165\linewidth}
        \centering
        \includegraphics[width=\linewidth]{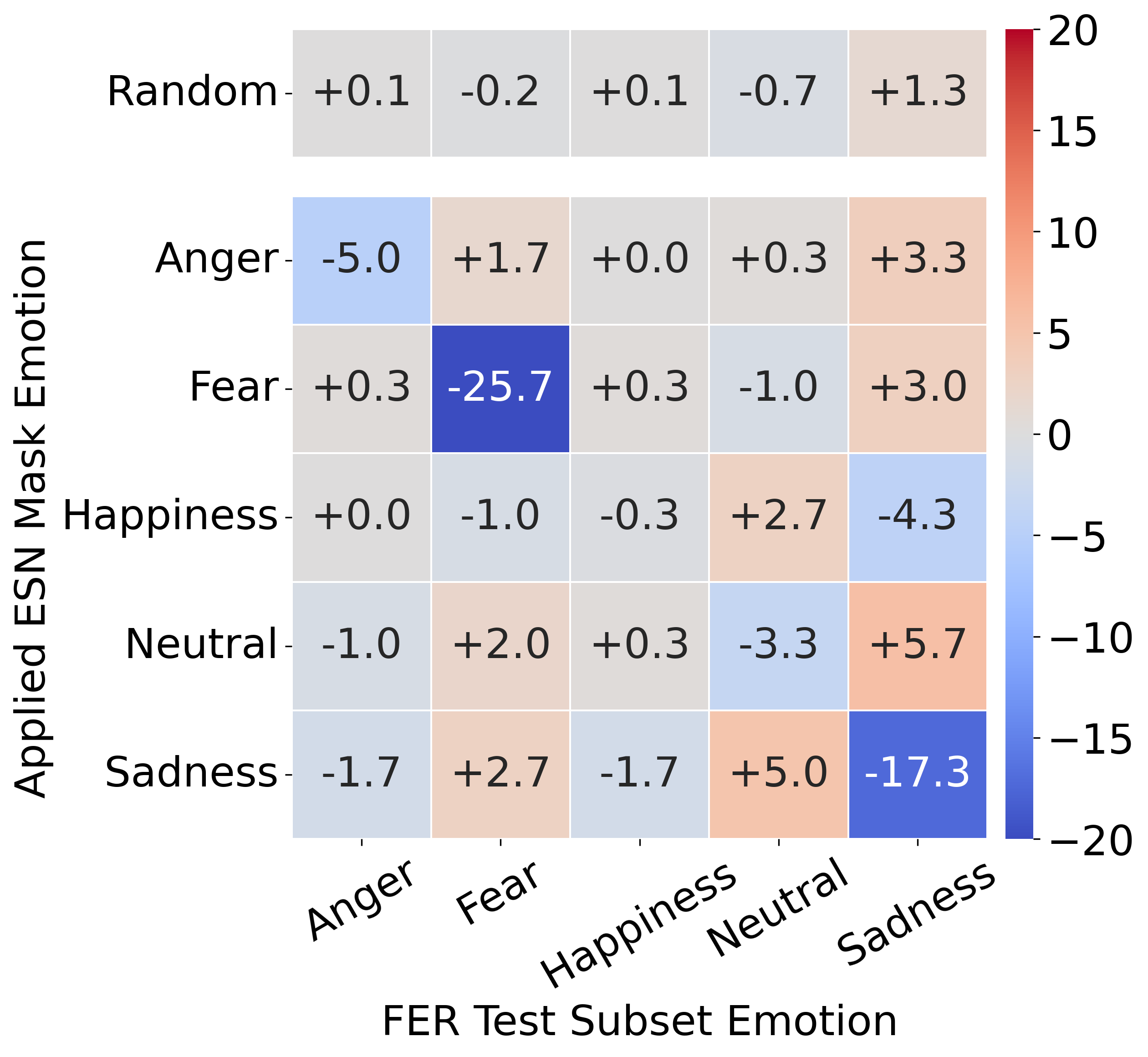}
        \caption{Deactivation,\\MiniCPM-o-4.5}
    \end{subfigure}\hfill
    \begin{subfigure}[b]{0.165\linewidth}
        \centering
        \includegraphics[width=\textwidth]{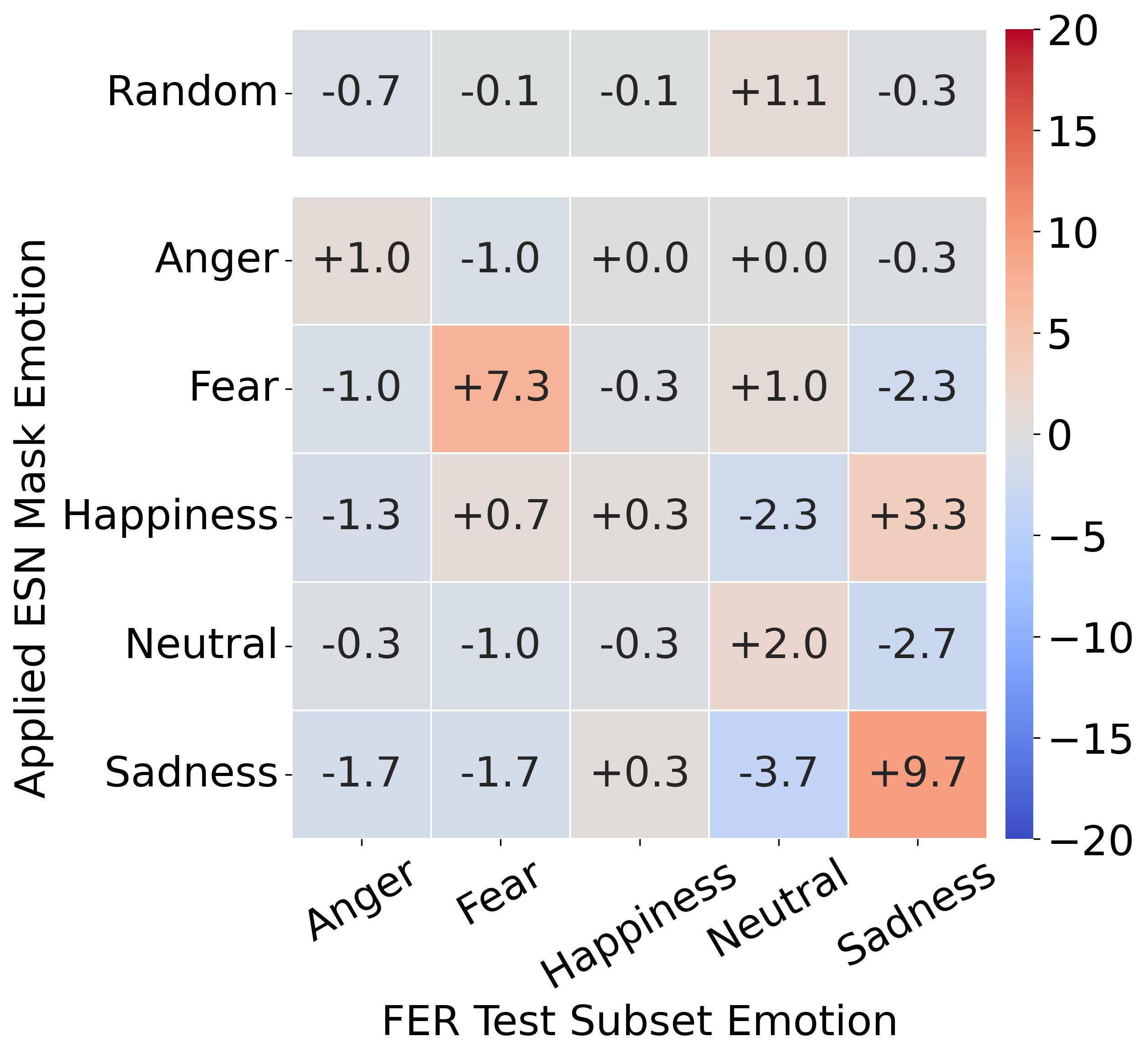}
        \caption{Steering,\\MiniCPM-o-4.5}
    \end{subfigure}\hfill
    \begin{subfigure}[b]{0.165\linewidth}
        \centering
        \includegraphics[width=\linewidth]{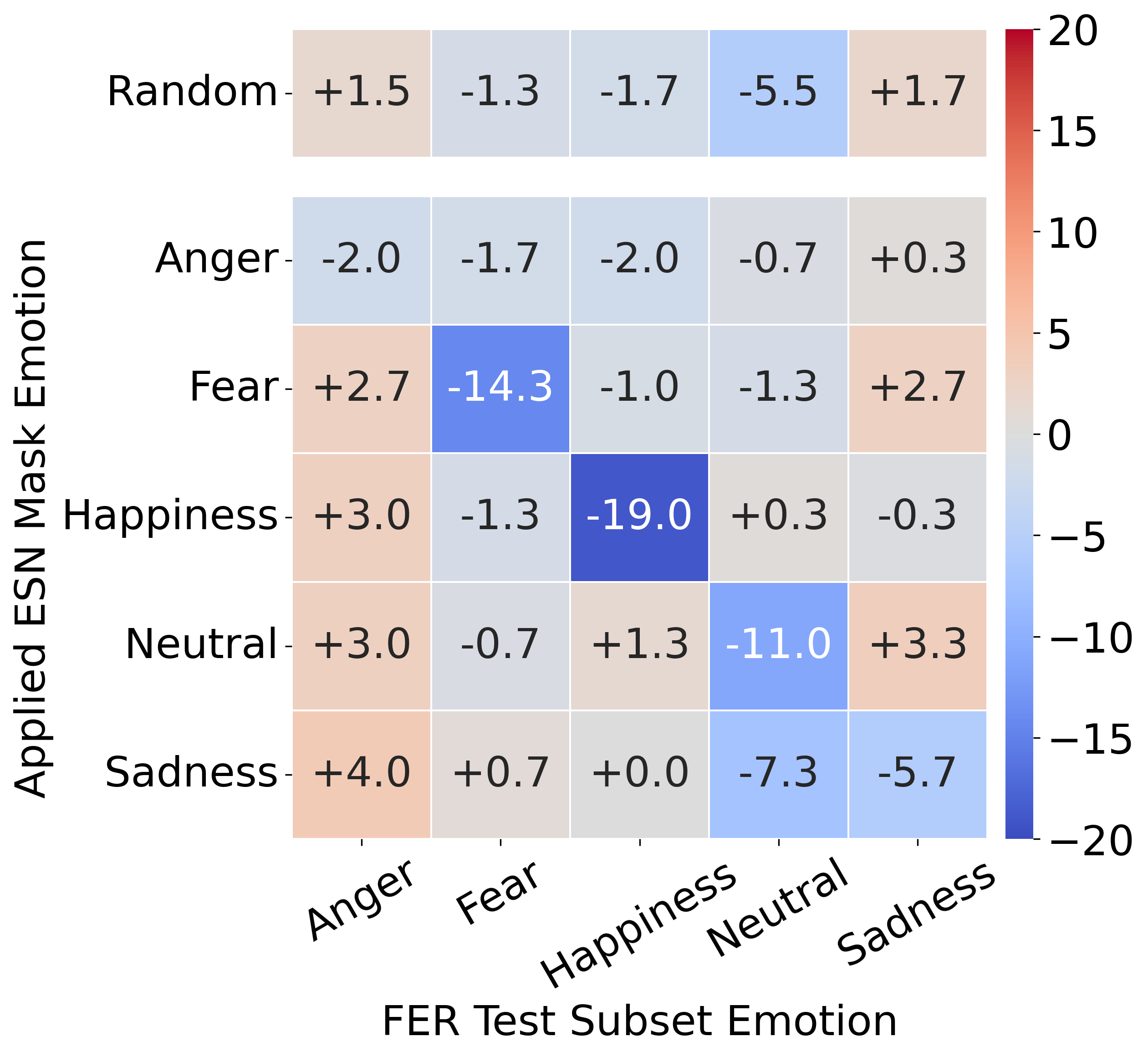}
        \caption{Deactivation,\\Qwen2.5-Omni-7B}
    \end{subfigure}\hfill
    \begin{subfigure}[b]{0.165\linewidth}
        \centering
        \includegraphics[width=\linewidth]{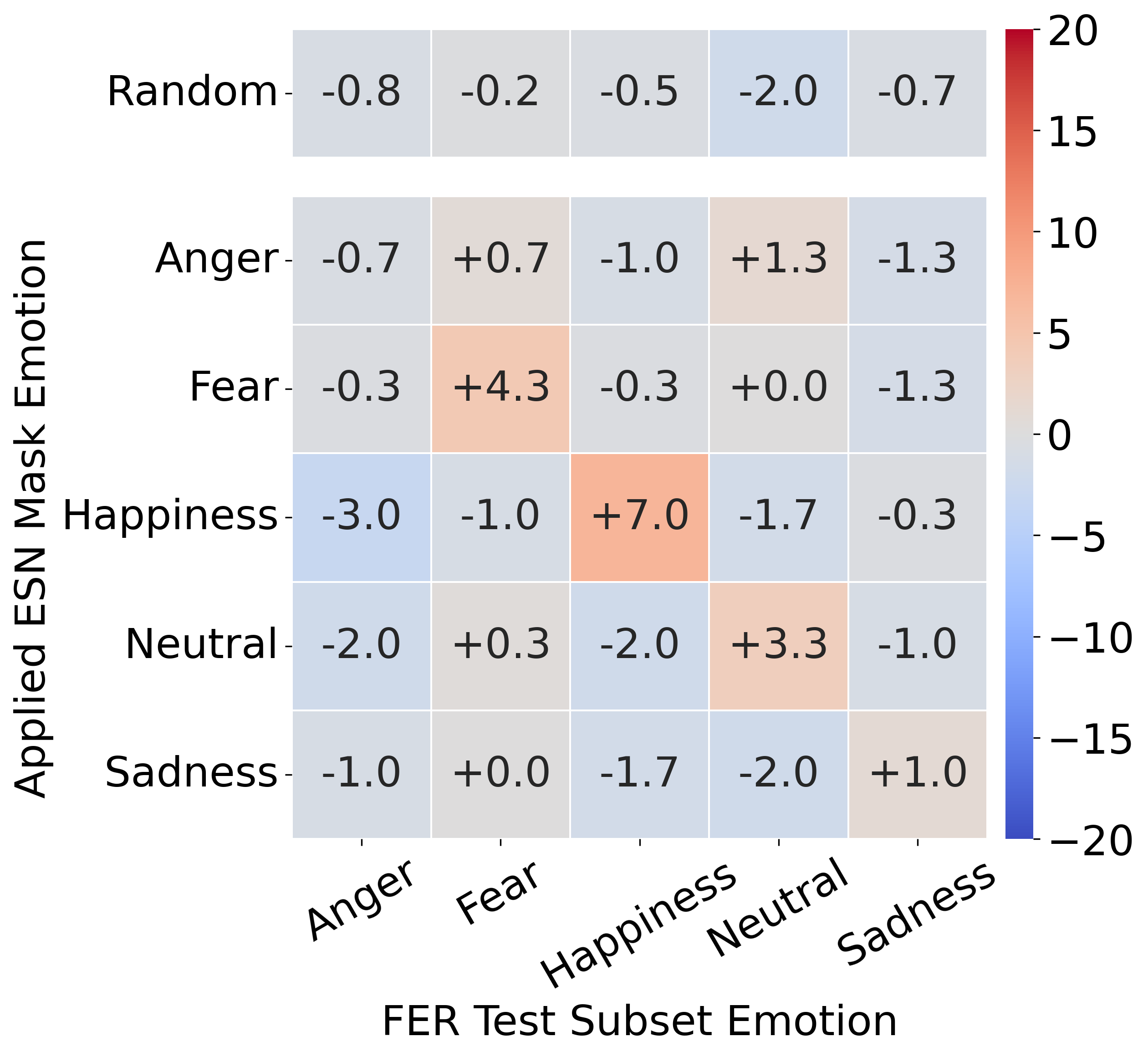}
        \caption{Steering,\\Qwen2.5-Omni-7B}
    \end{subfigure}\hfill
\caption{\textbf{Emotion-specific FER effects of V-ESN interventions.}
Heatmaps report changes in FER accuracy after applying FER-derived V-ESN masks.
Rows denote the mask emotion and columns denote the FER test-subset emotion; ``Random'' gives same-size random-mask controls.
Diagonal cells correspond to matched mask--test emotion pairs.}
    \label{fig:mono_results}
\end{figure*}

\subsection{Evaluation Metrics}
\label{sec:eval_metrics}
We report baseline emotion-recognition performance using unweighted average recall (UAR) over the five emotion categories. For intervention analysis, all effects are measured relative to the original, unintervened model under the same evaluation protocol. For an ESN mask associated with emotion $e$, the Self-Emotion score is the post-intervention accuracy on test examples whose ground-truth label is $e$. We report the average of this score over all target emotions $e \in E$. The Avg. Cross-Emotion score is computed by first averaging, for each target mask $e$, the post-intervention accuracies on all non-target emotion subsets $e' \neq e$, and then averaging over target emotions. The $\Delta$ values in the tables denote changes relative to the corresponding unmasked UAR under the same evaluation protocol. 
Under deactivation, a negative gap indicates selective degradation of matched-emotion recognition; under steering, a positive gap indicates selective improvement in matched-emotion recognition relative to non-target emotions.

\section{Results}
\label{sec:results}

\subsection{Mono-Modal Causal Validation of ESNs}
\label{sec:mono}

We first test whether ESNs discovered from a modality are causally involved in emotion recognition within that same modality.
As shown in Table~\ref{tab:mono_results}, this validation succeeds for both FER-derived V-ESNs and SER-derived A-ESNs across all three MFMs.
Deactivation consistently reduces performance on the matched emotion more than on non-matched emotions, yielding negative Self-Cross Gaps, while steering produces the opposite pattern with positive gaps.
Same-size random masks have only mild effects, indicating that the results are not explained by arbitrary perturbations of decoder MLP activations.

Figure~\ref{fig:mono_results} gives a finer-grained view of the FER results. The heatmaps show a broadly matched-emotion structure: deactivation often produces the largest or among the largest decreases on the diagonal, and steering often increases the corresponding target-emotion response more than most non-target responses. The pattern is clearest for several affective categories such as anger, fear, happiness, and sadness, whereas neutral shows a less uniformly targeted pattern.
This difference is plausible because neutral is closer to an absence of overt affect than to a strongly expressed emotion, and its recognition may depend more on suppressing evidence for other emotions than on activating a single positive affective pattern.

This confirms that ESNs can be identified from facial-expression activations, extending prior evidence that SER-derived A-ESNs are functional in SER \cite{zhao-etal-2026-discovering}. 
Moreover, the approximately opposite and emotion-selective effects of deactivation and steering support the interpretation that these sparse neuron sets are emotion-specific causal components, providing the mono-modal basis for the cross-modal analyses that follow.

\begin{figure*}[ht!]
    \centering
    \begin{subfigure}[b]{0.161\textwidth}
        \centering
        \includegraphics[width=\linewidth]{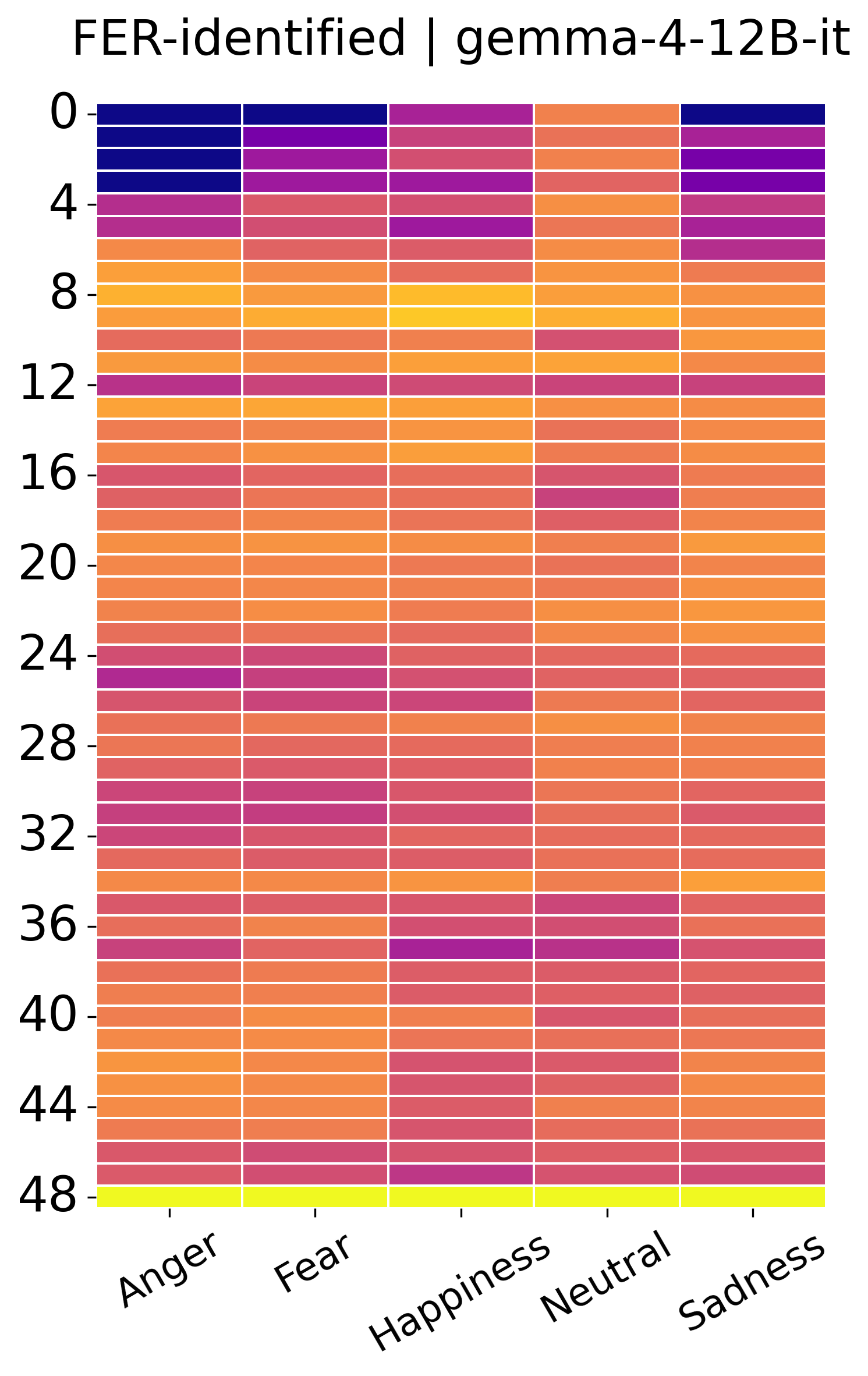}
    \end{subfigure}\hfill
    \begin{subfigure}[b]{0.161\textwidth}
        \centering
        \includegraphics[width=\linewidth]{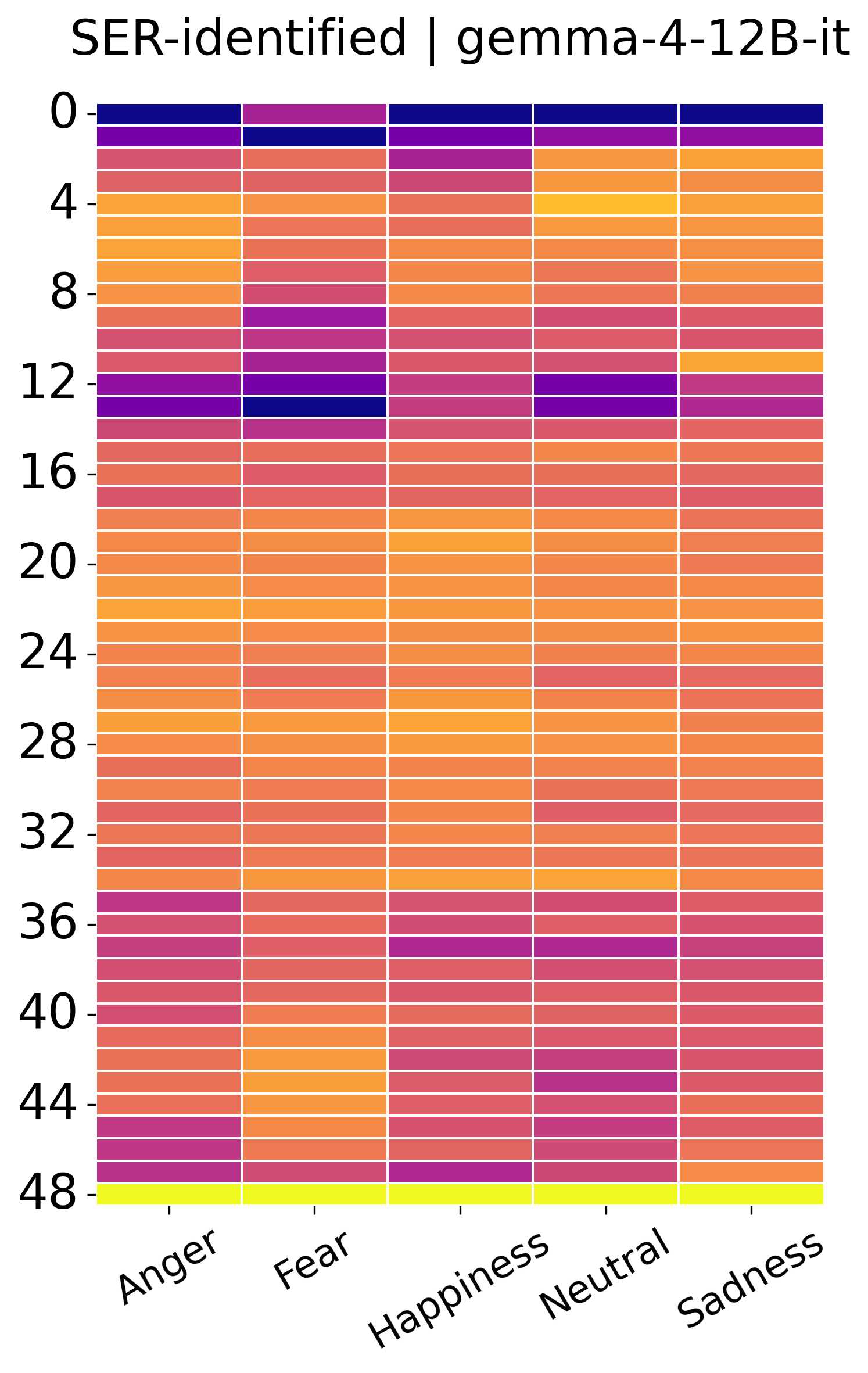}
    \end{subfigure}\hfill
    \begin{subfigure}[b]{0.157\textwidth}
        \centering
        \includegraphics[width=\linewidth]{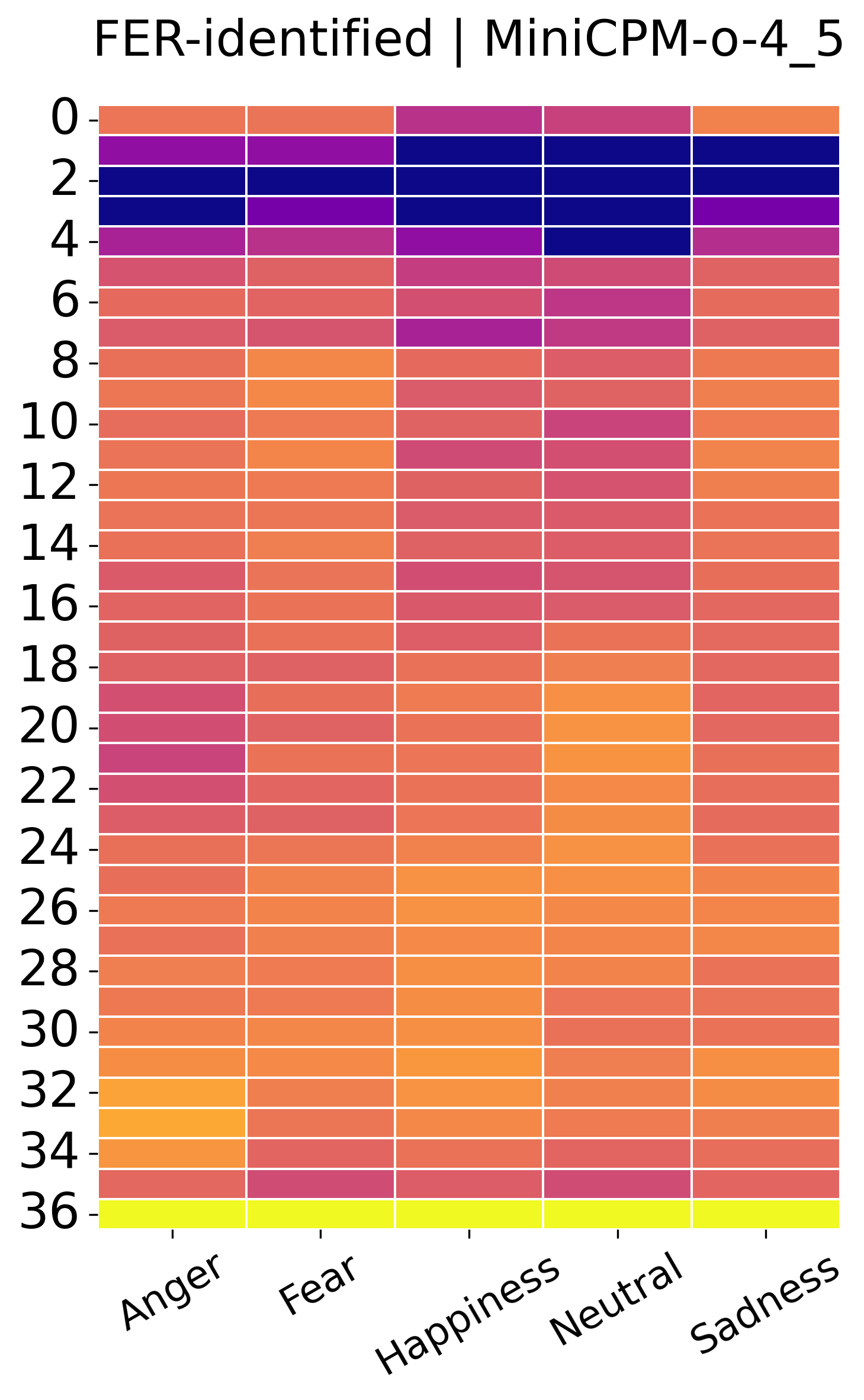}
    \end{subfigure}\hfill
    \begin{subfigure}[b]{0.157\textwidth}
        \centering
        \includegraphics[width=\linewidth]{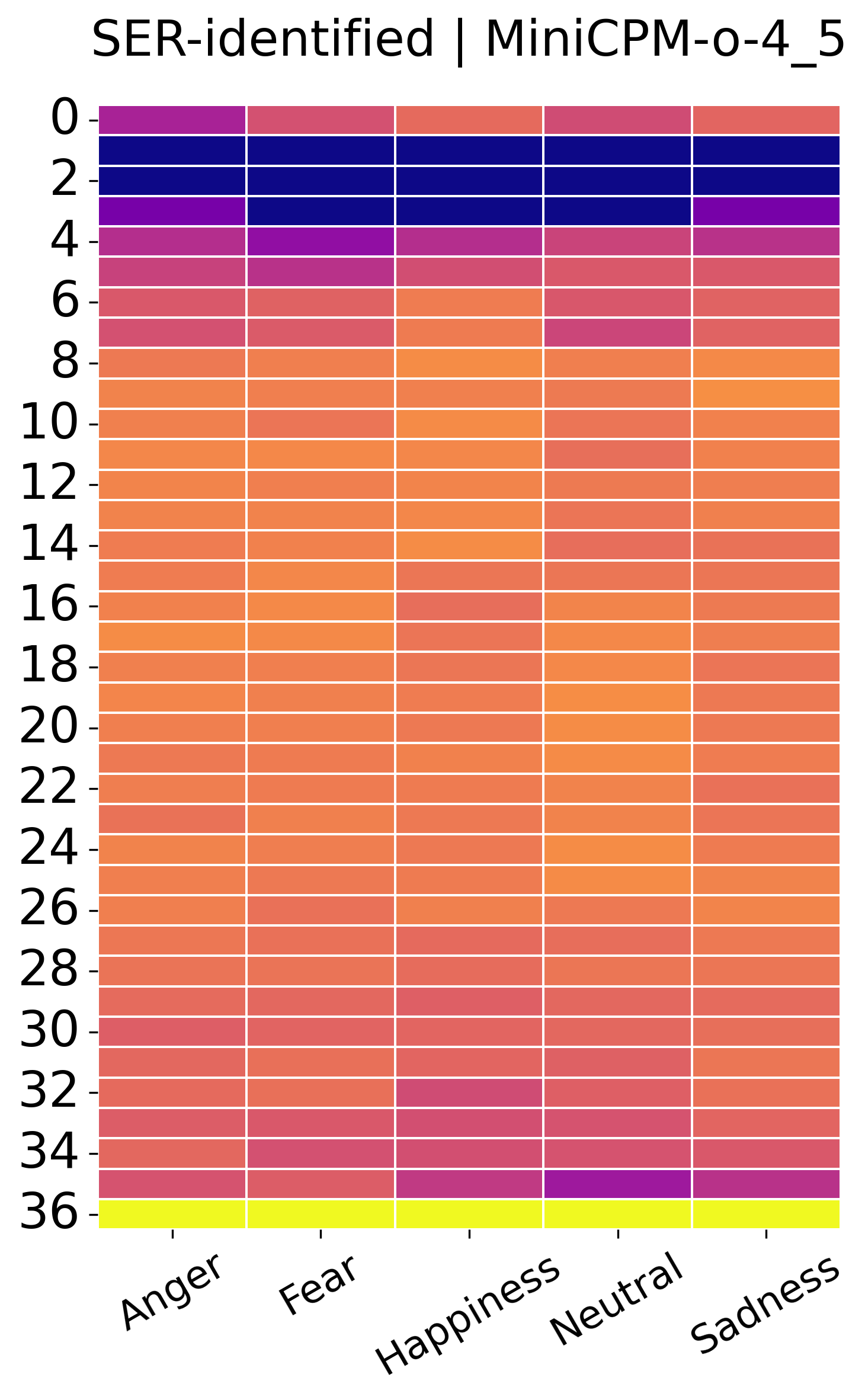}
    \end{subfigure}\hfill
    \begin{subfigure}[b]{0.165\textwidth}
        \centering
        \includegraphics[width=\linewidth]{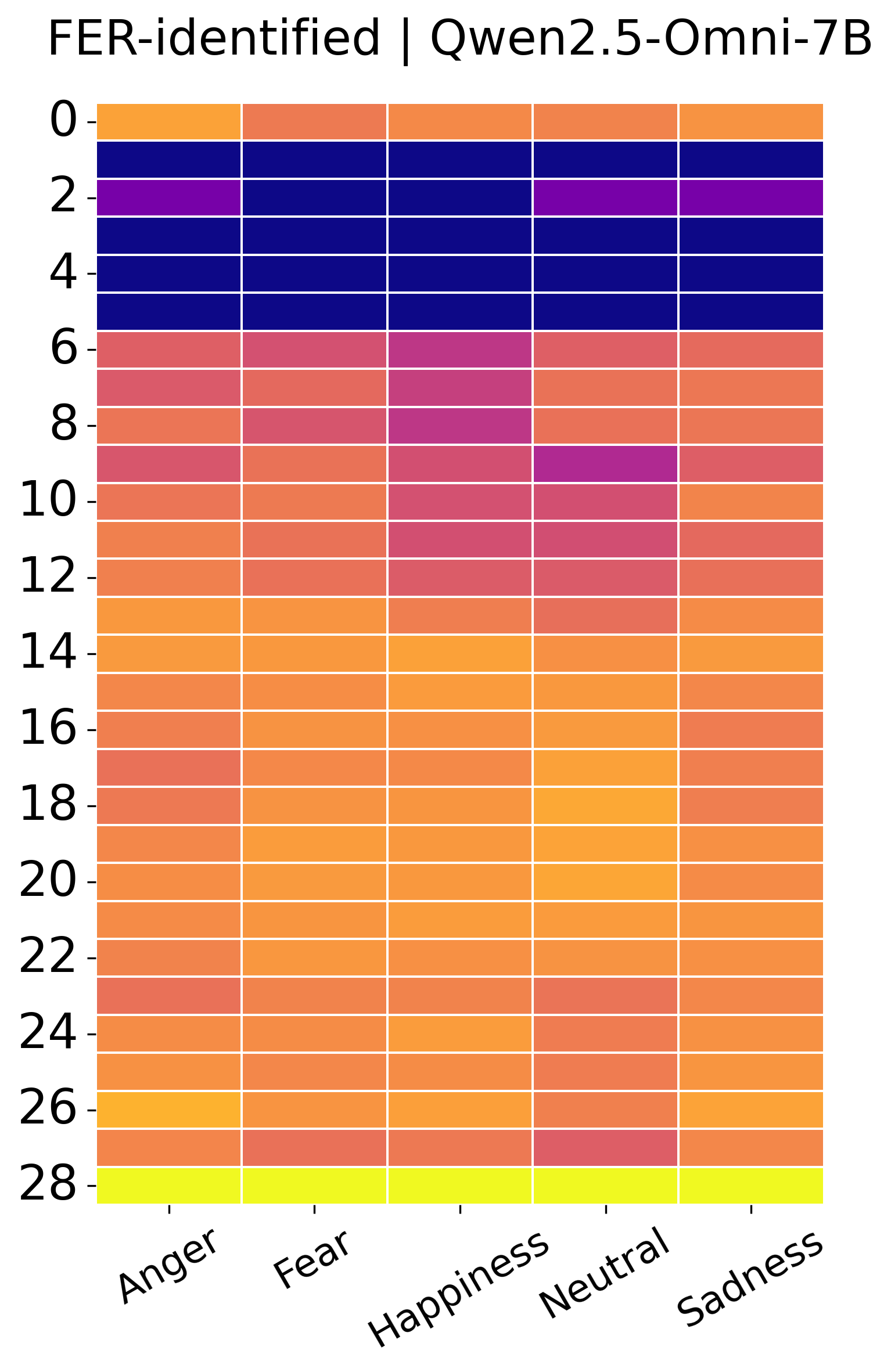}
    \end{subfigure}\hfill
    \begin{subfigure}[b]{0.195\textwidth}
        \centering
        \includegraphics[width=\linewidth]{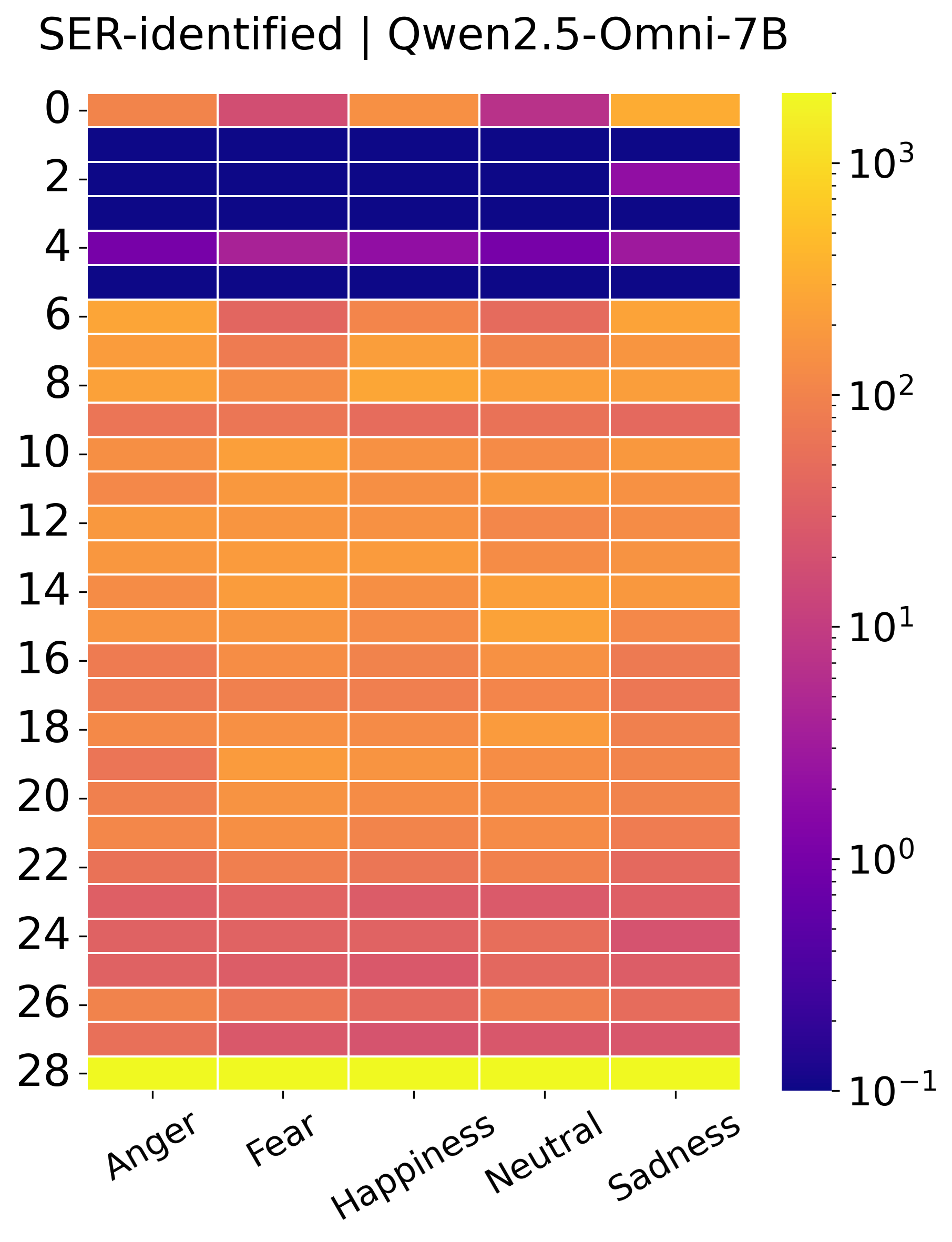}
    \end{subfigure}
    \caption{Layer-wise distribution of selected ESNs (ConAct-selected, $r=0.5\%$). Panels show FER-derived V-ESNs and SER-derived A-ESNs for each model; rows denote decoder MLP layers and columns denote emotions. Colors indicate neuron counts on a logarithmic scale. ESNs are sparse but span multiple layers, with stronger concentration in middle and later layers.}
    \label{fig:layers}
\end{figure*}

\begin{figure}[ht!]
    \centering
    \begin{subfigure}[b]{0.33\linewidth}
        \centering
        \includegraphics[width=\linewidth]{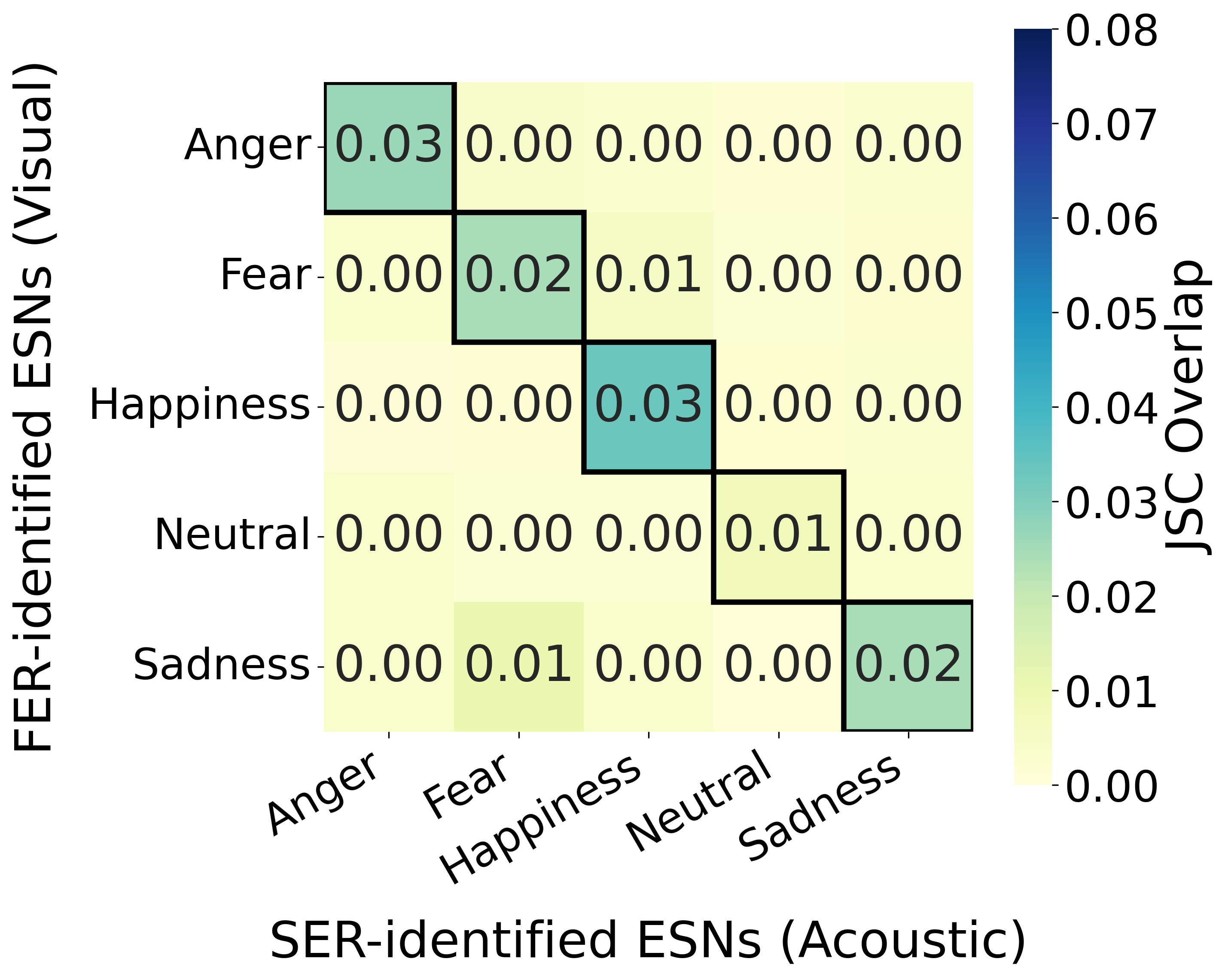}
        \caption{Gemma-4-12B-it}
    \end{subfigure}\hfill
    \begin{subfigure}[b]{0.33\linewidth}
        \centering
        \includegraphics[width=\linewidth]{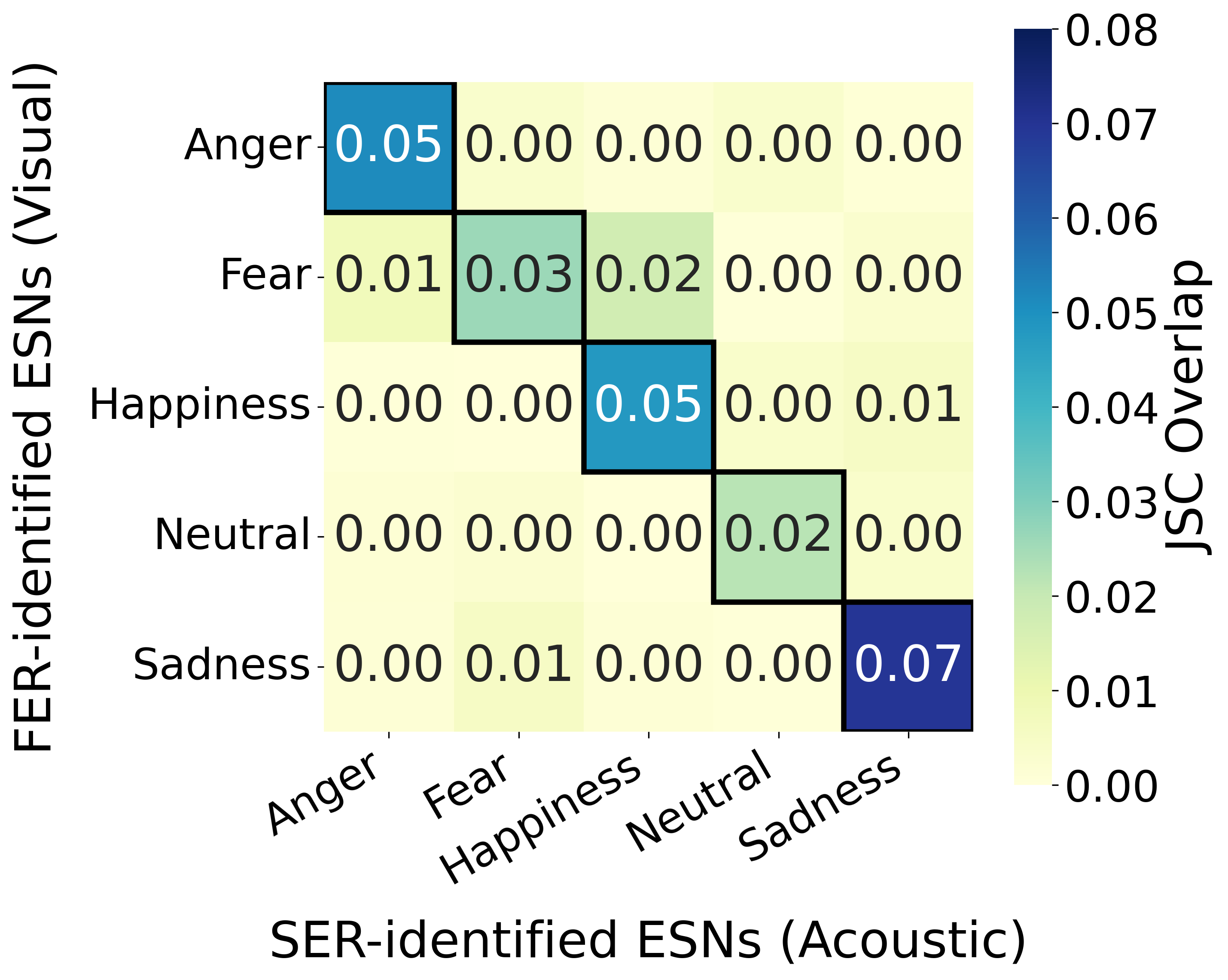}
        \caption{MiniCPM-o-4.5}
    \end{subfigure}\hfill
    \begin{subfigure}[b]{0.33\linewidth}
        \centering
        \includegraphics[width=\linewidth]{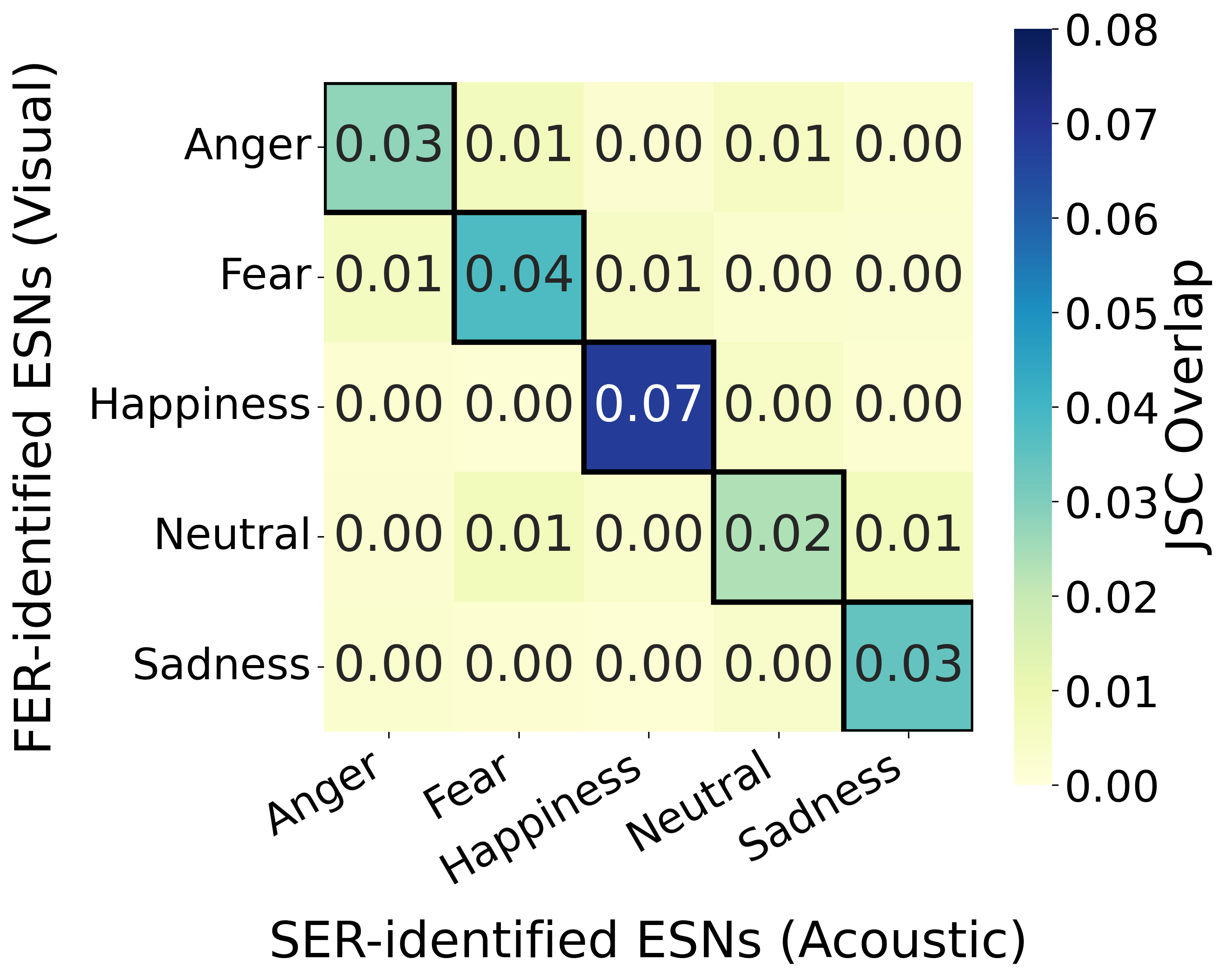}
        \caption{Qwen2.5-Omni-7B}
    \end{subfigure}\hfill
    \caption{\textbf{Cross-modal overlap between FER-derived V-ESNs and SER-derived A-ESNs.} Each heatmap reports Jaccard similarity coefficient across emotion-specific neuron sets; rows denote FER emotions, columns denote SER emotions, and darker cells indicate larger overlap.}
    \label{fig:crossmodal_overlap}
\end{figure}

\subsection{Acoustic and Visual ESNs Show Cross-Modal Alignment}
We next examine whether FER-derived V-ESNs and SER-derived A-ESNs are modality-specific or partially shared. We analyze both emotion-level neuron-set overlap and layer-wise ESN localization.
Figure~\ref{fig:crossmodal_overlap} reports the Jaccard similarity between A-ESNs and V-ESNs for each pair of emotion categories. Across all three models, the overlap is sparse but tends to be diagonal-dominant: neuron sets for the same speech and facial emotion overlap more than mismatched emotion pairs, whose similarities are often near zero. The strongest matched-category overlaps vary by model: Gemma-4-12B-it shows clearer alignment for anger, happiness, and sadness; MiniCPM-o-4.5 for sadness, with additional alignment for anger, happiness, and neutral; and Qwen2.5-Omni-7B for happiness and sadness. The absolute similarity coefficients are small, as expected from selecting only the top 0.5\% of decoder MLP neurons, but the matched-emotion structure suggests that the overlap is not merely due to generic task activation.

Figure~\ref{fig:layers} shows where the selected neurons occur across decoder depth. In all models, both A-ESNs and V-ESNs are distributed over multiple layers rather than confined to a single block. They are relatively rare in early layers and more frequent in middle and later layers, consistent with emotion information becoming more task- or decision-relevant deeper in the decoder. The FER and SER profiles are broadly similar within each model, although their exact localization differs: MiniCPM-o-4.5 has dense middle-to-late ESN selection for both modalities, Qwen2.5-Omni-7B shows a more concentrated intermediate band, and Gemma-4-12B-it is more diffuse across depth. The late-layer concentration also requires caution, because decoder states near the output may partly encode category-level decision variables or answer-selection processes. However, this pattern is unlikely to be explained solely by lexical association \cite{wang-etal-2025-audio}: the option order is randomized, models are instructed to output only the option index, and, more importantly, ESNs identified from different modalities and datasets exhibit emotion-matched overlap and emotion-specific causal effects, including under cross-modal transfer. We therefore interpret the late-layer ESNs as decoder-level affective functional units that may combine perceptual evidence with emotion-category decision representations, rather than as purely lexical-answer neurons.

Notably, these similarities arise despite A-ESNs and V-ESNs being identified from different tasks, datasets, and input modalities. This makes it less likely that the diagonal overlap and comparable layer-wise profiles are explained solely by the selection procedure. Instead, they suggest that speech and facial emotion recognition may converge onto a sparse set of decoder MLP units associated with emotion-relevant processing. These structural results suggest partial alignment between acoustic and visual ESNs, but they do not establish functional sharing. We therefore test whether ESNs selected in one modality causally affect recognition in the other.

\subsection{Cross-Modal ESN Interventions Reveal Bidirectional Transfer}

\begin{table*}[ht!]
\centering
\resizebox{0.95\textwidth}{!}{%
\begin{tabular}{@{}cccccccc@{}}
\toprule
Evaluation Setting & MFM      & Intervention                & Unmasked               & Random ($\Delta$) & Self-Emotion ($\Delta$) & Avg.\ Cross-Emotion ($\Delta$) & Self-Cross Gap \\ \midrule
                   & Gemma-4-  & Deactivation $\downarrow$ & \multirow{2}{*}{60.47} & 60.19 (-0.28)     & 42.40 (-18.07)          & 61.33 (+0.86)                 & -18.93         \\
                   & 12B-it  & Steering $\uparrow$       &                        & 60.13 (-0.34)     & 67.40 (+6.93)           & 59.12 (-1.35)                 & +8.28          \\ \cmidrule(l){2-8} 
\textbf{FER}       & MiniCPM-  & Deactivation $\downarrow$ & \multirow{2}{*}{64.00} & 64.13 (+0.13)     & 58.67 (-5.33)           & 63.73 (-0.27)                 & -5.06          \\
with A-ESN masks   & o-4.5   & Steering $\uparrow$       &                        & 63.99 (-0.01)     & 65.80 (+1.80)           & 63.72 (-0.28)                 & +2.08          \\ \cmidrule(l){2-8} 
                   & Qwen2.5-  & Deactivation $\downarrow$ & \multirow{2}{*}{48.60} & 47.53 (-1.07)     & 41.40 (-7.20)           & 47.87 (-0.73)                 & -6.47          \\
                   & Omni-7B & Steering $\uparrow$       &                        & 47.76 (-0.84)     & 51.80 (+3.20)           & 47.40 (-1.20)                 & +4.40          \\ \midrule
                   & Gemma-4-  & Deactivation $\downarrow$ & \multirow{2}{*}{35.87} & 35.47 (-0.40)     & 27.33 (-8.54)           & 36.70 (+0.83)                 & -9.37          \\
                   & 12B-it  & Steering $\uparrow$       &                        & 35.65 (-0.22)     & 39.87 (+4.00)           & 35.00 (-0.87)                 & +4.87          \\ \cmidrule(l){2-8} 
\textbf{SER}       & MiniCPM-  & Deactivation $\downarrow$ & \multirow{2}{*}{27.47} & 27.01 (-0.46)      & 23.33 (-4.14)           & 28.20 (+0.73)                 & -4.87          \\
with V-ESN masks   & o-4.5   & Steering $\uparrow$       &                        & 27.33 (-0.14)     & 30.53 (+3.06)           & 26.73 (-0.74)                 & +3.80          \\ \cmidrule(l){2-8} 
                   & Qwen2.5-  & Deactivation $\downarrow$ & \multirow{2}{*}{36.93} & 36.99 (+0.06)     & 29.60 (-7.33)           & 38.03 (+1.10)                 & -8.43          \\
                   & Omni-7B & Steering $\uparrow$       &                        & 36.83 (-0.10)     & 39.33 (+2.40)           & 35.97 (-0.96)                 & +3.36          \\ \bottomrule
\end{tabular}%
}
\caption{\textbf{Cross-modal causal effects of ESN interventions.}
SER-derived A-ESN masks are applied during FER, and FER-derived V-ESN masks are applied during SER. Columns follow Table~\ref{tab:mono_results}.}
\label{tab:cross_results}
\end{table*}

Having established that FER-derived V-ESNs are causally involved in FER, we next ask whether these neurons are modality-specific or whether they also participate in affective processing across modalities. To this end, we perform cross-modal interventions: SER-derived A-ESN masks are applied during FER, and FER-derived V-ESN masks are applied during SER. In both settings, the neuron sets are identified in one modality but intervened on in the other, providing a direct test of whether ESNs discovered from one perceptual channel causally influence recognition behavior in another.

Table~\ref{tab:cross_results} summarizes the cross-modal intervention results. Across all three models, ESNs identified in one modality produce structured effects when transferred to the other modality, whereas random interventions generally remain close to the unmasked baseline. When applying A-ESNs to FER, deactivation reduces matched self-emotion performance for all models, with larger drops than the average changes on non-target emotions. Steering A-ESNs during FER produces the complementary pattern, increasing the corresponding self-emotion score and yielding positive Self-Cross Gaps. The magnitudes vary across models, with the strongest transfer observed for Gemma-4-12B-it and more modest but consistent effects for MiniCPM-o-4.5 and Qwen2.5-Omni-7B.

The reverse transfer shows the same qualitative pattern. When applying V-ESNs to SER, deactivation reduces recognition of the corresponding self-emotion across models, while random deactivation remains close to zero. Steering V-ESNs improves SER performance on the corresponding self-emotion category, producing positive Self-Cross Gaps for all models. These results suggest that FER-derived V-ESNs can participate in SER behavior, although the effect sizes indicate partial rather than complete sharing between the modalities. The bidirectional transfer reduces the likelihood that the effect is specific to a single modality, and instead suggests a recurring organization in which decoder MLPs contain sparse emotion-sensitive components accessible from both speech and facial cues.

\begin{figure}[ht!]
    \centering
    \begin{subfigure}[b]{0.35\linewidth}
        \centering
        \includegraphics[width=\linewidth]{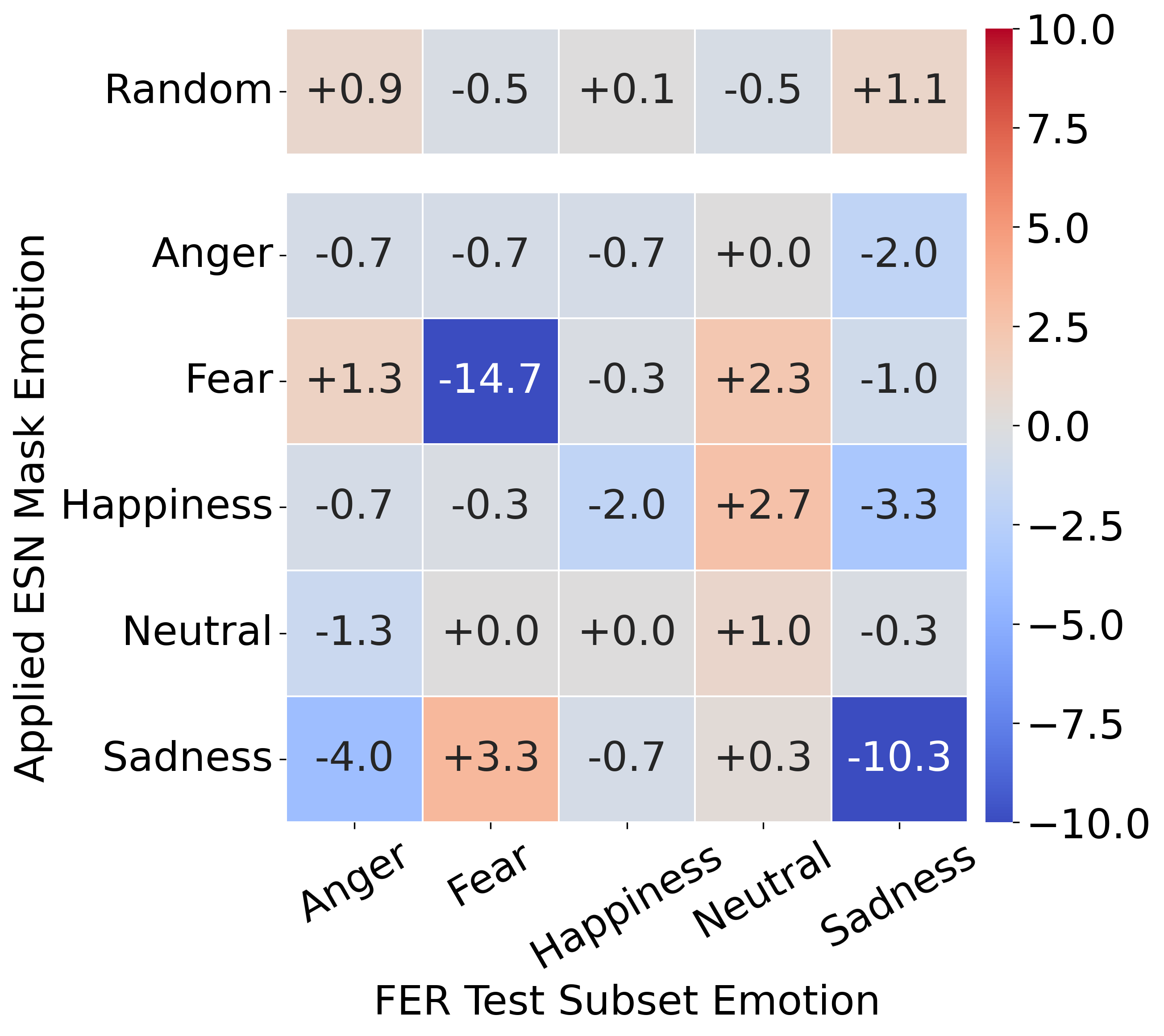}
        \caption{Deactivation, A-ESN on FER}
    \end{subfigure}
    \begin{subfigure}[b]{0.35\linewidth}
        \centering
        \includegraphics[width=\linewidth]{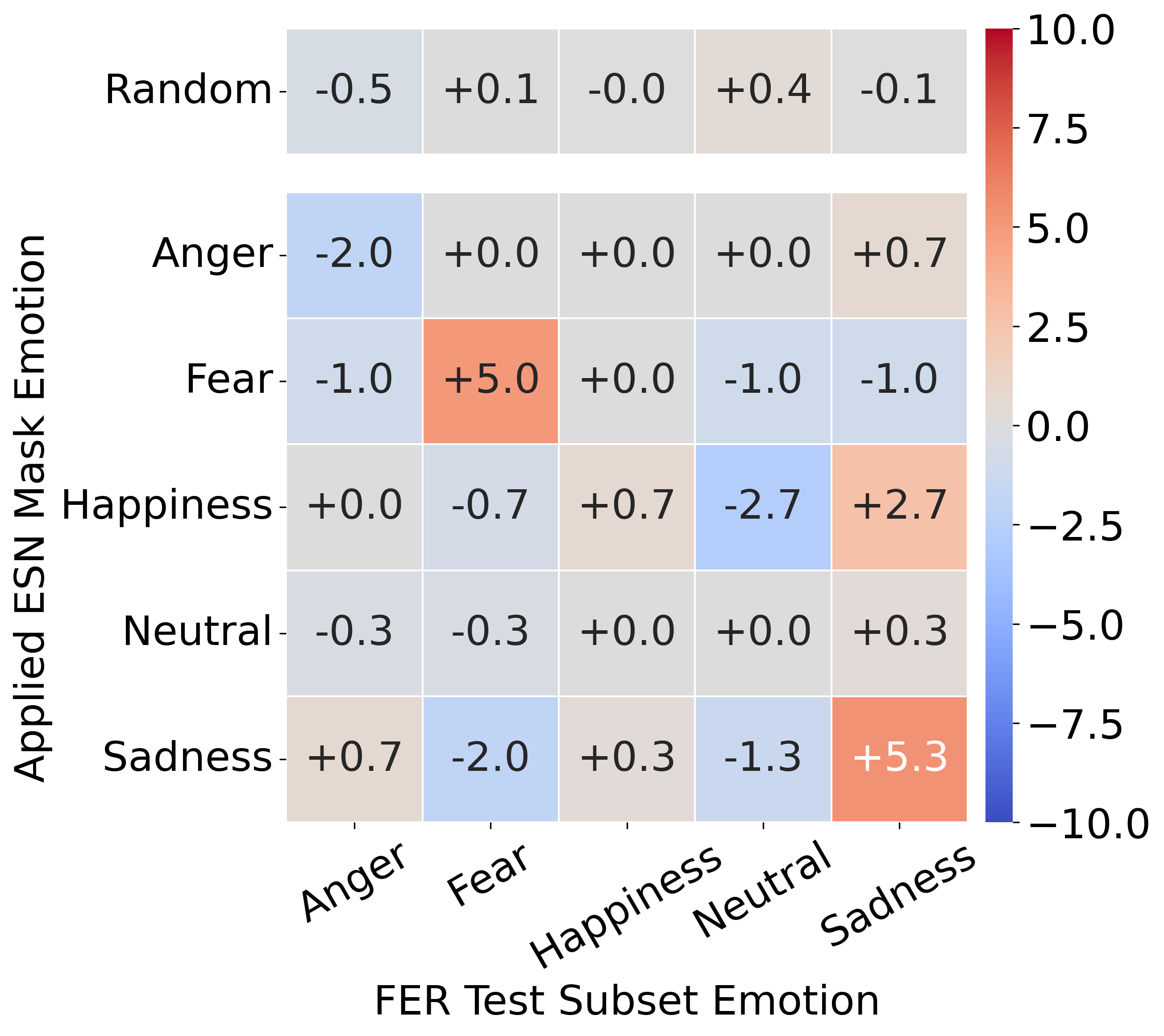}
        \caption{Steering, A-ESN on FER}
    \end{subfigure}\hfill

    \vspace{0.5em}

    \begin{subfigure}[b]{0.35\linewidth}
        \centering
        \includegraphics[width=\linewidth]{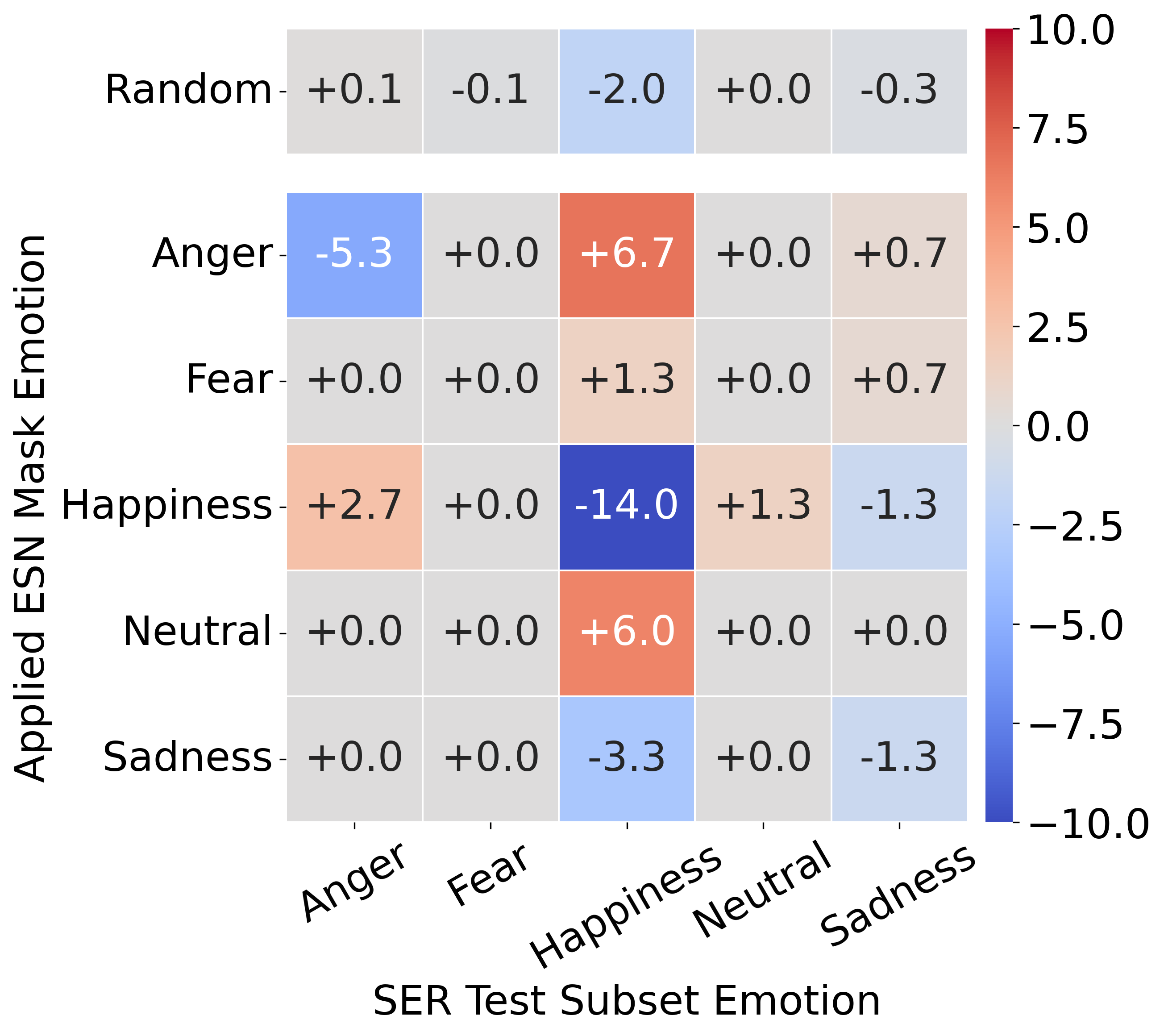}
        \caption{Deactivation, V-ESN on SER}
    \end{subfigure}
    \begin{subfigure}[b]{0.35\linewidth}
        \centering
        \includegraphics[width=\linewidth]{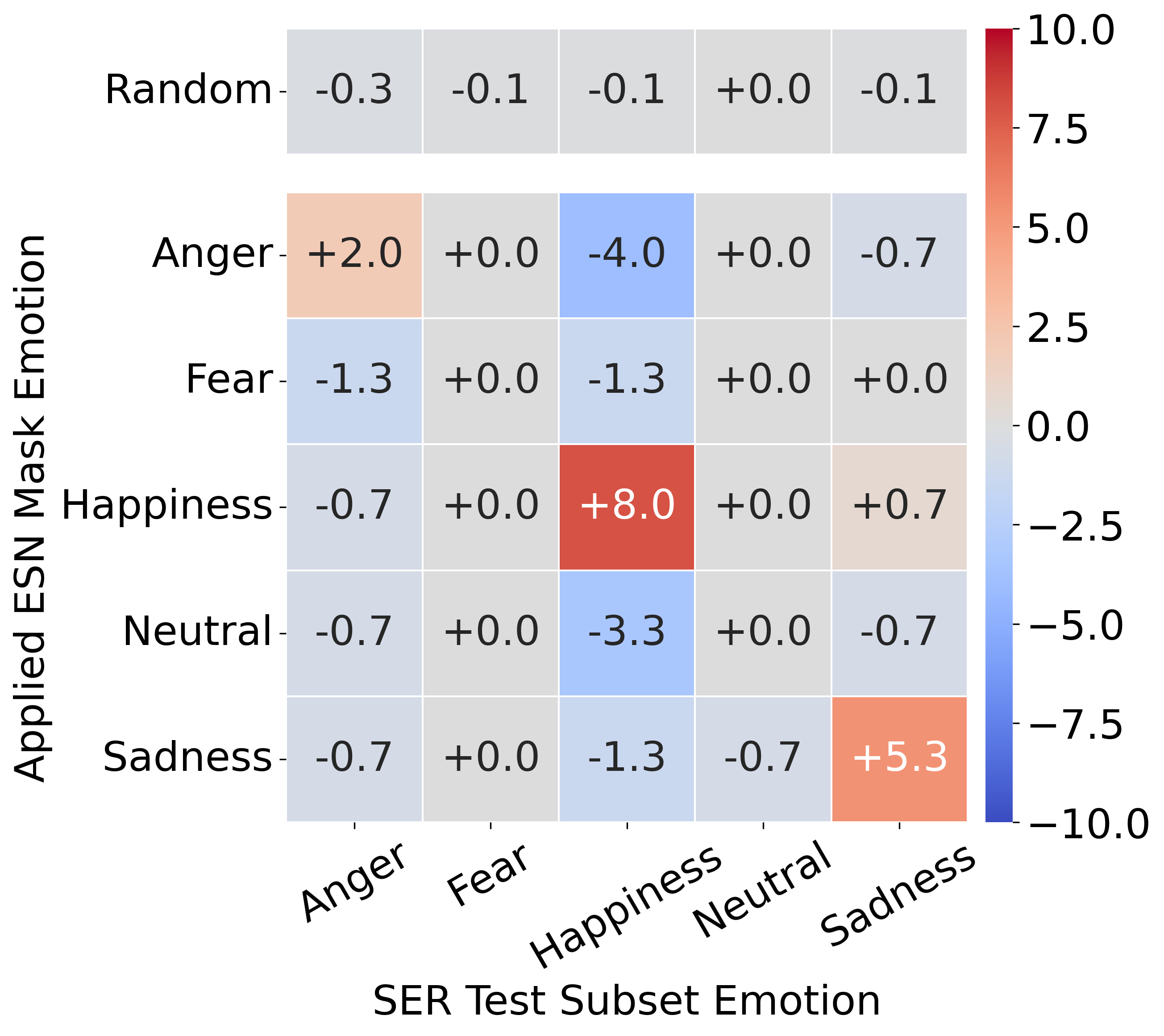}
        \caption{Steering, V-ESN on SER}
    \end{subfigure}
    \caption{\textbf{Emotion-specific effects of cross-modal ESN transfer for MiniCPM-o-4.5.} Heatmaps report changes in recognition accuracy when ESN masks identified from one modality are applied to the other modality. Panels (a,b) apply SER-derived A-ESN masks during FER, and panels (c,d) apply FER-derived V-ESN masks during SER.}
    \label{fig:cross_results}
\end{figure}

Figure~\ref{fig:cross_results} illustrates the emotion-specific intervention patterns for MiniCPM-o-4.5. The transferred interventions show a clearer matched-emotion structure in aggregate than the random controls, although the heatmaps also contain non-negligible off-diagonal effects. In particular, deactivation of transferred ESNs tends to reduce performance more for the corresponding self-emotion than for the average non-target emotions, while steering tends to increase the corresponding self-emotion response. These trends are consistent with the positive and negative Self-Cross Gaps in Table~\ref{tab:cross_results}, but they also indicate that cross-modal transfer is selective rather than perfectly emotion-isolated. Overall, these patterns strengthen the descriptive overlap analysis by showing that ESNs selected in one modality are not merely artifacts of the selection procedure, but can also causally modulate emotion-recognition behavior in the other modality.

\section{Conclusion}
\label{sec:conclusoin}

We investigated whether speech and facial emotion recognition in MFMs rely on separate modality-specific pathways or partially shared internal functional units. 
Across Gemma-4-12B-it, MiniCPM-o-4.5, and Qwen2.5-Omni-7B, we identified sparse ESNs from decoder MLP activations using SER and FER as complementary probes. 
Both acoustic ESNs and visual ESNs are causally meaningful within their respective modalities: deactivation selectively impairs recognition of the matched emotion, while steering selectively improves recognition of that emotion.
Beyond mono-modal validation, we found that A-ESNs and V-ESNs exhibit emotion-matched overlap and broadly comparable layer-wise distributions, with selected neurons tending to appear more in middle and later decoder layers. 
Most importantly, cross-modal interventions reveal partial bidirectional causal transfer: A-ESNs identified from speech affect facial emotion recognition, and V-ESNs identified from faces affect SER, with matched self-emotion effects that are stronger than the corresponding random-mask controls. 
Together, these results suggest that speech and facial affect processing in MFMs partially converges onto sparse decoder-level components that can be localized and manipulated without training.

This work provides one of the first cross-modality activation-level analyses of affective functional units in MFMs, especially across speech and facial emotion recognition. 
Our results offer a mechanistic view of how MFMs internally organize affective information across modalities. 
Future work could move beyond post-hoc localization toward building controllable affective interfaces for MFMs: identifying whether shared ESNs can be used to calibrate, debias, or personalize emotion perception across speech and vision.

\section*{Acknowledgment}
This work is supported by the National Science Foundation (NSF) CAREER Award IIS-2533652.

\bibliographystyle{IEEEtran}
\bibliography{preprint,custom,models,datasets}

\appendix

\section{Reproducibility}
\label{appendix:reproducibility}

\subsection{Datasets and Models}
\begin{table}[h]
\centering
\resizebox{0.5\linewidth}{!}{%
\begin{tabular}{@{}lcc@{}}
\toprule
Task      & SER         & FER       \\ \midrule
Dataset  & MSP-Podcast & AffectNet \\ \midrule
Anger     & 36260       & 25382     \\
Fear      & 1943        & 6878      \\
Happiness & 58684       & 134915    \\
Neutral   & 79117       & 75374     \\
Sadness   & 24629       & 25259     \\ \bottomrule
\end{tabular}%
}
\caption{Dataset statistics showing utterance/images counts per emotion.
}
\label{table:datasets}
\end{table}

\begin{table}[H]
\centering
\resizebox{0.48\textwidth}{!}{%
\begin{tabular}{@{}lll@{}}
\toprule
\textbf{Models} & \textbf{Hugging Face Identifier}                                            & \textbf{License}                \\ \midrule
Gemma-4-12B-it & \url{https://huggingface.co/google/Gemma-4-12B-it}              & Apache License 2.0              \\
MiniCPM-o-4.5   & \url{https://huggingface.co/openbmb/MiniCPM-o-4_5}            & Apache License 2.0              \\
Qwen2.5-Omni-7B & \url{https://huggingface.co/Qwen/Qwen2.5-Omni-7B}              & Apache License 2.0              \\ \bottomrule
\end{tabular}%
}
\caption{
Sources and licenses for the three evaluated MFMs.
}
\label{table:models}
\end{table}

\end{document}